%% file: main.tex
\documentclass[10pt,twocolumn,letterpaper]{article}

\usepackage[pagenumbers]{cvpr} 
\usepackage{comment}


\definecolor{cvprblue}{rgb}{0.21,0.49,0.74}
\usepackage[pagebackref,breaklinks,colorlinks,allcolors=cvprblue]{hyperref}

\usepackage{booktabs,multirow,graphicx}
\usepackage{array}
\usepackage[table]{xcolor}
\usepackage{arydshln}
\usepackage{colortbl}
\usepackage{pifont}   
\usepackage{amssymb}  
\usepackage{arydshln}
\usepackage[most]{tcolorbox}

\newcommand{\cmark}{\ding{51}}       
\newcommand{\xmark}{\ding{55}}       
\newcommand{\tmark}{\(\triangle\)}   

\def\paperID{14593} 
\def\confName{CVPR}
\def\confYear{2026}

\title{Text-Printed Image: Bridging the Image-Text Modality Gap for Text-centric Training of Large Vision-Language Models}

\author{Shojiro Yamabe$^{1,2,\dag}$ \quad Futa Waseda$^{1,3}$ \quad Daiki Shiono$^{1,4}$ \quad Tsubasa Takahashi$^{1,\ddag}$\\
$^1$Turing Inc. \quad $^2$Institute of Science Tokyo \quad $^{3}$The University of Tokyo \quad $^{4}$Tohoku University\\
{\tt\small $^{\dag}$yamabe.s.2fb0@m.isct.ac.jp \quad $^{\ddag}$tsubasa.takahashi@turing-motors.com}
}

\begin{document}


\twocolumn[{%
\renewcommand\twocolumn[1][]{#1}%
\maketitle
\begin{center}
    \centering
    \includegraphics[width=0.96\linewidth]{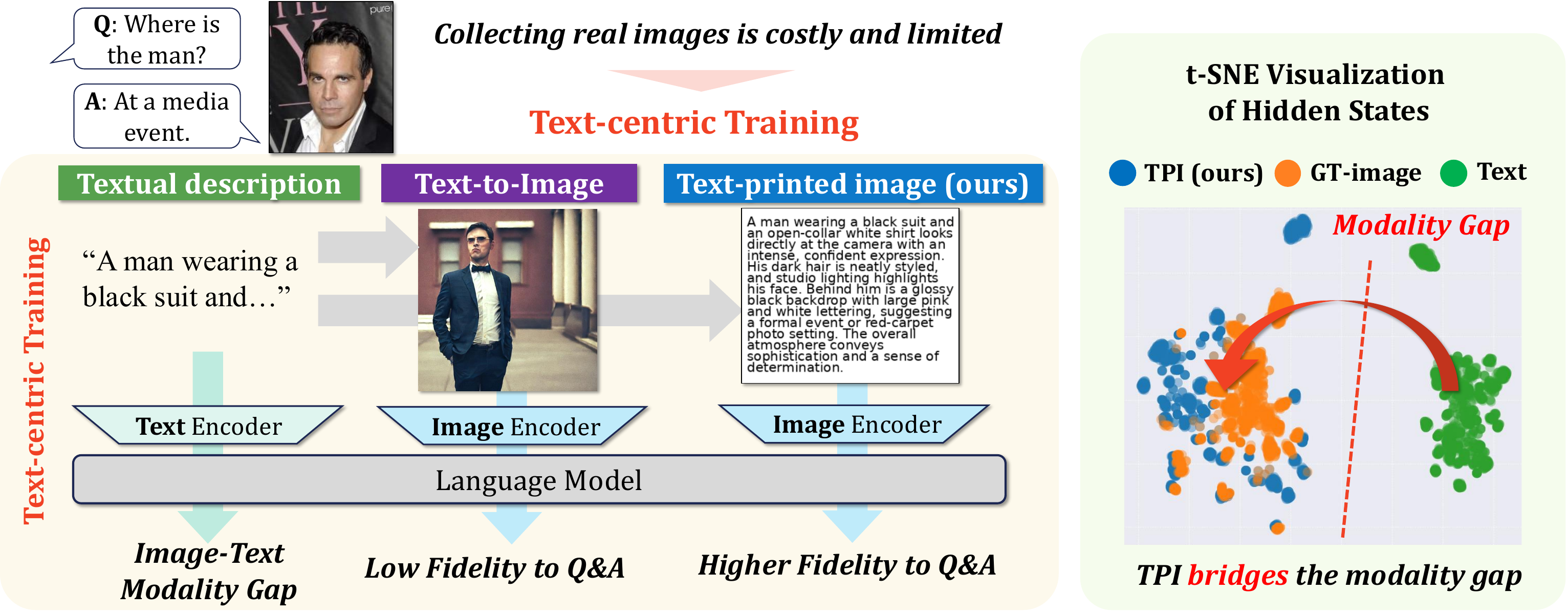}
    \captionof{figure}{\textbf{Text-Printed Image (TPI) provides an efficient and broadly applicable approach for text-centric training.} While raw text input suffers from an image-text modality gap and synthetic images generated by a text-to-image model lose fidelity to Q\&A pairs, TPI bridges this gap by embedding textual content into the visual pathway, achieving high fidelity to the ground-truth visual supervision.
    \label{fig:overview_figure}
    }
\end{center}}]

\input{sec/0_abstract}
\input{sec/1_intro}
\input{sec/2_related_works}

\input{sec/3_method}

\input{sec/4_experiment}
\input{sec/5_augmentation}

\input{sec/6_discussion}
\input{sec/7_conclusion}

\newpage

{
    \small
    \bibliographystyle{ieeenat_fullname}
    \bibliography{main}
}

\input{sec/X_suppl}

\end{document}

%% file: sec/0_abstract.tex
\begin{abstract}
Recent large vision–language models (LVLMs) have been applied to diverse VQA tasks.
However, achieving practical performance typically requires task-specific fine-tuning with large numbers of image-text pairs, which are costly to collect.
In this work, we study text-centric training, a setting where only textual descriptions are available and no real images are provided, as a paradigm for low-cost data scaling.
Unlike images, whose collection is often restricted by privacy constraints and scarcity in niche domains, text is widely available.
Moreover, text is easily editable, enabling automatic diversification and expansion with LLMs at minimal human effort.
While this offers clear advantages over image collection in terms of scalability and cost, training on raw text without images still yields limited gains on VQA tasks because of the image–text modality gap.
To address this issue, we propose a \textbf{Text-Printed Image (TPI)}, which generates synthetic images by directly rendering the given textual description on a plain white canvas.
This simple rendering projects text into the image modality and can be integrated into arbitrary existing LVLM training pipelines at low cost.
Moreover, TPI preserves the semantics of the text, whereas text-to-image models often fail to do.
Across four models and seven benchmarks, our systematic experiments show that TPI enables more effective text-centric training than synthetic images generated by a diffusion model.
We further explore TPI as a low-cost data-augmentation strategy and demonstrate its practical utility.
Overall, our findings highlight the significant potential of text-centric training and, more broadly, chart a path toward fully automated data generation for LVLMs.
\end{abstract}


%% file: sec/1_intro.tex
\section{Introduction}\label{sec:intro}

Large Vision-Language Models (LVLMs)~\cite{liu2023visual} have emerged as a powerful extension of Large Language Models (LLMs). Building on strong foundational capabilities, they have been applied across diverse VQA tasks, including autonomous driving~\cite{zhou2024vision} and healthcare~\cite{hartsock2024vision}. To attain practical performance on these downstream tasks, task-specific fine-tuning is typically required.

However, constructing a large-scale dataset remains challenging for such training, as practical SFT typically requires vast amounts of image–text instructions~\cite{liu2023visual, xu2023multiinstruct, xu2024vision}.
In contrast to LLM training, which benefits from abundant text corpora~\cite{brown2020language}, LVLM training requires image-conditioned instruction data, and constructing such data is substantially more costly~\cite{li2024visionlanguage, liu2023visual}.
Although automated methods for scraping web images and pairing them with corresponding text have been proposed~\cite{li2023llavamed, zhang2023llavar, deng2024enhancing, wang2024vigc, dai2025captions}, their applicability is often limited in specialized or niche domains~\cite{van2024large, li2023clip}.

In this work, we consider \emph{text-centric training}, where only textual descriptions are available and no images are provided. 
Our premise is that text is abundant and easy to prepare compared to images.
Rich text corpora and structured lexical resources make it possible to synthesize descriptions from class names and concept taxonomies~\cite{fan2024scaling, hammoud2024synthclip, wu2025synthetic}.
Moreover, text is much easier to modify than images.
While changing a small detail in an image is difficult~\cite{hertz2023prompttoprompt}, textual descriptions can be edited or expanded with little effort, and LLMs can automatically generate diverse variants~\cite{wei2019eda}.
Recent advances in LLM-based data augmentation~\cite{wang2023self, ding2024data} further enhance this potential, enabling scalable and diverse text synthesis at low cost.
Taken together, these factors make text-centric training a promising yet underexplored direction for advancing LVLMs when images are scarce or unavailable.

A primary challenge for effective text-centric training is the \emph{image-text modality gap}~\cite{liang2022mind, zhang2023diagnosing, garrido2023on,zhang2024connect}, which is a mismatch between the latent distributions of textual and visual representations within a model. Because of this gap, training only on given raw text fails to generalize effectively to visual inputs, as the representations learned from text differ from those required to process images during inference.

A straightforward approach to bridge this gap is to synthesize images using text-to-image models (T2I), such as diffusion models~\cite{podell2024sdxl}; however, T2I often struggles to maintain fidelity to the given textual descriptions~\cite{hu2023tifa, huang2023t2i}.
The generated images often deviate from the intended semantic content, making them unsuitable as training data. 
Indeed, our experimental results confirm that T2I-generated images are poorly relevant to their corresponding queries and responses, leading to suboptimal performance. Mitigating this issue typically requires extensive sampling and manual verification, which substantially increases costs.


To bridge the modality gap while maintaining semantic fidelity, we propose a low-cost yet effective method, \emph{Text-Printed Image (TPI)}.
Given a textual description, TPI renders the text onto a plain white canvas and uses the resulting image as the visual input during training.
This design ensures that the LVLM receives the full semantics of the text while the image encoder still processes it in the visual modality, thereby preserving semantic fidelity and mitigating the modality gap.
In our experiments, TPI-based training outperforms training on T2I-generated images and matches or approaches the performance achieved with ground-truth images on several datasets.
Furthermore, TPI can be applied without a GPU, and it is architecture-agnostic because it requires no changes to the training pipeline.
Overall, the combination of low cost and strong effectiveness makes TPI a highly practical approach for LVLM training.

Finally, we introduce a data augmentation approach based on TPI. We automatically generate synthetic textual descriptions by using an LLM and evaluate models trained on this data.
Even when seeding the augmentation with only 1\% of the original dataset, we can observe measurable performance gains.
Moreover, augmenting the full training set with the synthetic text yields additional gains over the model trained only on the full dataset. These results reinforce that text-centric training with TPI is a promising direction for low-cost automatic training data generation for LVLMs.

\vspace{-12pt}
\paragraph{Contributions.}
Our contributions are summarized as:
\begin{itemize}
\item We introduce a light-weight and architecture-agnostic text-centric training method, \emph{Text-Printed Image (TPI)}, that bridges the modality gap while preserving text semantics by rendering textual descriptions as images.
\item We conduct extensive evaluations over multiple LVLMs and downstream tasks, showing that training with TPI can outperform training with T2I-generated images and, on some benchmarks, match or closely approach training with ground-truth images.
\item We investigate a practical text-driven augmentation that leverages LLMs to generate instruction data, showing that TPI enables performance improvements even with only 1\% of the original dataset.
\end{itemize}

%% file: sec/2_related_works.tex
\section{Related works}

\subsection{Large Vision-Language Models}
LVLMs are a family of models built on LLMs that interpret not only language but also visual information~\cite{liu2023visual, dai2023instructblip, zhu2024minigpt, lin2024vila, bai2023qwen, wang2024qwen2, chen2024internvl, zhu2025internvl3, wang2025internvl3, tong2024cambrian}. 
These models are typically trained on paired image–text data. 
While these models exhibit strong general-purpose capabilities, they typically require large amounts of training data. 
In this work, we focus on LVLMs with this architecture and study text-centric training to mitigate data scarcity.

\subsection{Text-centric training}
\paragraph{CLIP-based approach.}
Prior work has explored text-centric training and proposed methods to reduce the modality gap~\cite{liang2022mind, zhang2023diagnosing, li2023decap, zhou2022lafite2, nukrai2022text, zhou2023shifted, zhang2024connect, sharifzadeh2024synth, yaras2025explaining}.
However, they are designed for CLIP-style dual encoders and do not transfer directly to LVLMs.
Specifically, while they assume a one-to-one match between image and text representation dimensions, this assumption does not hold for LVLMs because text representations are token-based and their dimensionality changes with the length of the input text.

\vspace{-8pt}
\paragraph{T2I-based approach.}
Several approaches use T2I to generate synthetic images from given descriptions~\cite{aboutalebi2024magid, ashrafian2024vision, wu2024autohallusion, ben2024mitigating, li2024enhanced, wu2025less, gui2025cycleaug, trabucco2024effective, wu2025synthetic, fan2024scaling}. However, as demonstrated in our experiments, T2I-generated images struggle to reflect the given text faithfully, and training on them does not yield sufficient performance improvements. Moreover, obtaining high-quality images requires substantial trial-and-error and quality checks, making the process costly.

\vspace{-8pt}
\paragraph{Text-centric training for LVLMs.}
Recent works focus on text-centric training for LVLMs~\cite{yu2025unicorn, hu2025praxis, choi2024improving, du2025virgo}.
Some of them~\cite{hu2025praxis, du2025virgo} showed that training LVLMs solely on text, analogous to LLM training, can improve their reasoning ability.
Although applications to VQA tasks have been explored~\cite{choi2024improving}, the gains remain limited due to the image–text modality gap, as our experiments later demonstrate.
Yu et al.~\cite{yu2025unicorn} proposed to mitigate this gap by adding a constant vector to the text features.
However, similar to CLIP-based approaches, their method assumes that image and text features have the same dimensionality, which makes it difficult to apply to many existing LVLMs. In addition, the constant vector must be added not only during training but also at inference time, incurring extra inference cost.
In this work, we propose an architecture-agnostic and low-cost training method for VQA tasks that mitigates this gap without any inference-time modifications.

\subsection{Other synthetic data generation approaches}
To enable scalable training, many studies have explored synthetic data generation that is not text-centric.
A common strategy leverages existing image sources~\cite{lee2021constructing, li2023llavamed, zhang2023llavar, deng2024enhancing, lee2024dialogcc, wang2024vigc, dai2025captions, zhao2025omnialign, chen2024allava, zhao2024genixer, irawan2025towards, wang2025endochat, zhou2025anyprefer}.
These methods construct synthetic datasets by collecting images from the web or other corpora and attaching high-quality captions.
However, they face an inherent limitation: applicability diminishes when publicly available images are scarce, particularly in specialized domains.
Although this line of work is important, our work focuses on text-centric training. Consequently, methods that rely on sourcing images to synthesize datasets fall outside the scope of this paper.

%% file: sec/3_method.tex
\section{Method}\label{sec:method}
We propose TPI for effective text-centric training, where only textual descriptions are available.
Preparing text is easier than images: Text is abundant and can be assembled from concept resources such as taxonomies or class names~\cite{fan2024scaling, hammoud2024synthclip, wu2025synthetic}, and diverse variants can be generated automatically with LLMs at minimal human cost~\cite{wang2023self, ding2024data}.
However, effective training remains challenging due to the image–text modality gap.
To address this issue, we investigate how to learn effectively from given textual descriptions and introduce a method that enables effective training.

\subsection{Problem Formulation}\label{subsec:problem_formulation}
\noindent
\textbf{Notation.}
Recent LVLMs, such as LLaVA~\cite{liu2023visual} and Qwen-VL~\cite{bai2023qwen, wang2024qwen2}, typically consist of a vision encoder, a projector, and an LLM.
Given an image–query pair $(i, q)$, the model divides the image into patches and projects them into image features $v = p_{\theta}(i)$, where $p_{\theta}(\cdot)$ denotes the image processor composed of the vision encoder and the projector.
The image features are then fed into the LLM $f_{\phi}(\cdot)$, resulting in the conditional distribution for response $r$:
\begin{equation}
    f_{\phi}\!\left(r \mid p_\theta(i), q \right).
\end{equation}

\noindent
\textbf{Text-centric training.}
In text-centric training, the image $i$ is unavailable; instead, only a textual description $t$ is given.
We introduce a transformation $T$ that synthesizes features directly from text.
Given a training dataset $\mathcal{D}_{\mathrm{txt}}$ consisting of triplets $(t, q, r)$, the objective is formulated as:
\begin{equation}\label{eq:sft}
\mathcal{L}(\theta,\phi)
=\mathbb{E}_{(t,q,r)\sim \mathcal{D}_{\mathrm{txt}}}
\!\left[-\log f_{\phi}\!\left(r \mid T(t), q\right)\right].
\end{equation}
For instance, if $T$ is the text encoder of the LLM, the model is trained using raw text in the same way as a standard LLM.
Alternatively, if $T$ is composed of a T2I generator $G$ and the image processor, such as $T(t) = p_\theta(G(t))$, it means that the training uses synthetic images created by the generator $G$ from text $t$.

\subsection{Difficulty and Requirements}\label{subsec:requirements}

The main difficulty in effective text-centric training is the \emph{image–text modality gap}, an inherent bias of the vision encoder whereby image features projected by the vision encoder and text features lie in systematically different regions of the representation space~\cite{liang2022mind, zhang2023diagnosing, garrido2023on,zhang2024connect, yaras2025explaining}.
As a result, even when an image and its textual description express the same semantics, the LVLM can interpret them as disparate signals, so the effectiveness of learning from raw text alone remains limited.
Thus, it is desirable that the synthesized feature $s=T(t)$ aligns with the image feature $v_t=f_{\theta}(i_t)$, where $i_t$ denotes a hypothetical image that visually represents the textual description $t$.

In addition, to design a practical solution, we require the transformation $T$ to satisfy the following three criteria:
\vspace{-12pt}
\paragraph{(R1) Compatibility with pretrained LVLMs:}
To ensure applicability to a broad range of LVLMs, the transformation $T$ should not rely on assumptions about model architecture or the visual encoding pipeline.
CLIP-based methods~\cite{liang2022mind, zhang2024connect, sharifzadeh2024synth, yaras2025explaining, yu2025unicorn} that assume alignment between text and image feature dimensions cannot be applied to most LVLMs because their text feature dimensions vary with the input length.
Thus, alternative architecture-agnostic approaches are needed for LVLMs.
\vspace{-12pt}
\paragraph{(R2) Preservation of text semantics:}
To maintain data quality, the synthesized feature $s = T(t)$ should faithfully preserve the semantics of the given text $t$.
T2I-generated images often fail to faithfully follow the provided text, leading to inconsistencies with the paired query and response.
\vspace{-12pt}
\paragraph{(R3) Efficiency and scalability:}
To preserve the low-cost advantage of text-centric training, $T$ should be computationally efficient and scalable. Additional training of $T$ or reliance on expensive text-to-image synthesis pipelines limit scalability. T2I are resource-intensive and often require domain-specific fine-tuning, undermining this efficiency advantage.

\subsection{Text-Printed Images}\label{subsec:text-printed_images}

To address the difficulty and satisfy the above requirements, we introduce \emph{TPI} that renders the given textual description directly onto a plain white canvas (Figure~\ref{fig:overview_figure}).
Our key insight is that, by directly rendering the provided text, TPI is processed by the vision encoder before reaching the LLM, while preserving its complete semantics.
The routing is crucial: the modality gap originates from biases and mismatches induced by the image encoder, so addressing this issue requires leveraging representations obtained from the image encoder.
At the same time, TPI preserves the full semantics of the description since the image explicitly contains the original text. It satisfies the requirement of semantic fidelity (R2).

Formally, let $R(\cdot;\psi)$ be a deterministic renderer that maps text to an RGB image with layout parameters $\psi$, such as font size and image size. The transformation is:
\begin{equation}
    T_{\text{print}}(c_i) = p_{\theta}(R(c_i;\psi)).
\end{equation}

We summarize how TPI compares with previous approaches against the above requirements in Table~\ref{tab:method_compare} and find that TPI satisfies (R1–R3). 
For \textbf{(R1)}, TPI uses the unmodified visual pathway: the synthesized image is processed by the existing image encoder and projector, without retraining of the projector or LLM, yielding a true drop-in replacement in standard training code.
For \textbf{(R2)}, as mentioned above, a TPI is generated by directly rendering the textual description $t$ onto an image, thus faithfully preserving its semantics. We empirically validate this in Section~\ref{subsubsec:relevance_score}.
For \textbf{(R3)}, TPI relies solely on a deterministic, training-free renderer $R$ and can be executed extremely efficiently, maintaining a clear cost advantage over text-to-image baselines, achieving roughly three orders of magnitude higher throughput as shown in Section~\ref{subsubsec:efficiency}.
Together, TPI provides a practical and scalable way to realize effective text-centric training for existing LVLMs.

\begin{table}[t]
\centering
\caption{Comparison of text-centric training methods.}
\label{tab:method_compare}
\vspace{-2mm}
\footnotesize
\setlength{\tabcolsep}{6pt}
\renewcommand{\arraystretch}{1.2}
\resizebox{0.50\textwidth}{!}{%
\begin{tabular}{lcccc}
\toprule
\textbf{Method} & \textbf{\shortstack[c]{Modality Gap\\Reduction}}  & \textbf{\shortstack[c]{(R1)\\Compatibility}} & \textbf{\shortstack[c]{(R2)\\Semantic fidelity}} & \textbf{\shortstack[c]{(R3)\\Efficiency}}\\
\midrule
Text-only & & \cmark & \cmark & \cmark \\
CLIP-based & \cmark & & \cmark & \cmark \\
Text-to-Image & \cmark & \cmark & &  \\
\rowcolor{gray!15} \textbf{TPI (ours)} & \cmark & \cmark & \cmark & \cmark \\
\bottomrule
\end{tabular}
}
\vspace{-4mm}
\end{table}

%% file: sec/4_experiment.tex
\begin{table*}[ht]
\centering
\caption{\textbf{TPI (ours) overcomes the image-text modality gap and achieves superior performance.} \emph{Text-only} offers only limited gains due to the gap. Each value represents the test score obtained after fine-tuning the pretrained model on the training set of each task.}
\vspace{-2mm}
\label{tab:vlm_benchmark}
\setlength{\tabcolsep}{4pt}
\renewcommand{\arraystretch}{1.1}
\footnotesize
\resizebox{0.98\linewidth}{!}{
\begin{tabular}{l l *{7}{r} r}
\toprule
& & \multicolumn{7}{c}{Tasks} & \\ 
\cmidrule(lr){3-9}
Model & Training & ScienceQA & OK-VQA & VizWiz & ChartQA & InfoVQA & DocVQA & DriveLM & Avg. \\
\midrule
\multirow{5}{*}{\rotatebox[origin=c]{90}{LLaVA 7B}}
& Pretrained                   & 66.09 & 49.87 & 52.89 & 17.12 & 21.91 & 23.86 & 47.84 & 39.94 \\
\cdashline{2-10}[4.0pt/2.5pt]
& Text-only                 & 72.63 & 60.73 & 59.25 & 19.24 & 25.36 & 28.61 & 64.03 & 47.12 \\
& Text-to-Image                    & 75.01 & 61.16 & \textbf{64.14} & 18.88 & 26.02 & 25.18 & \textbf{68.61} & 48.43 \\
& \textbf{Text-printed Image (Ours)}       & \textbf{75.11} & \textbf{61.70} & 61.96 & \textbf{23.28} & \textbf{28.43} & \textbf{33.80} & 65.49 & \textbf{49.97} \\
\cdashline{2-10}[4.0pt/2.5pt]
& \cellcolor{gray!15} GT-Image (Oracle) & \cellcolor{gray!15} 78.78 & \cellcolor{gray!15} 62.12 & \cellcolor{gray!15} 67.56 & \cellcolor{gray!15} 36.68 & \cellcolor{gray!15} 29.81 & \cellcolor{gray!15} 39.93 & \cellcolor{gray!15} 74.20 & \cellcolor{gray!15} 55.58 \\
\midrule
\midrule
\multirow{5}{*}{\rotatebox[origin=c]{90}{LLaVA 13B}}
& Pretrained                   & 71.39 & 52.49 & 56.33 & 19.32 & 25.87 & 27.86 & 51.93 & 43.60 \\
\cdashline{2-10}[4.0pt/2.5pt]
& Text-only                   & 75.61 & 64.59 & 63.79 & 24.32 & 29.21 & 32.85 & 64.50 & 50.70 \\
& Text-to-Image                    & 76.15 & 63.84 & \textbf{64.67} & 21.04 & 29.76 & 27.80 & \textbf{69.62} & 50.41 \\
& \textbf{Text-printed Image (Ours)}     & \textbf{76.30} & \textbf{64.73} & 64.46 & \textbf{28.72} & \textbf{30.75} & \textbf{36.70} & 64.00 & \textbf{52.24} \\
\cdashline{2-10}[4.0pt/2.5pt]
& \cellcolor{gray!15} GT-Image (Oracle) & \cellcolor{gray!15} 80.12 & \cellcolor{gray!15} 64.53 & \cellcolor{gray!15} 66.96 & \cellcolor{gray!15} 40.28 & \cellcolor{gray!15} 33.73 & \cellcolor{gray!15} 43.07 & \cellcolor{gray!15} 74.86 & \cellcolor{gray!15} 57.65 \\
\midrule
\midrule
\multirow{5}{*}{\rotatebox[origin=c]{90}{Qwen2.5 VL}}
& Pretrained                   & 76.65 & 43.88 & 70.80 & 83.28 & 80.21 & 94.36 & 39.49 & 69.81 \\
\cdashline{2-10}[4.0pt/2.5pt]
& Text-only                   & 86.53 & 61.78 & 68.54 & 85.92 & 77.60 & 92.89 & 66.26 & 77.07 \\
& Text-to-Image                    & 88.20 & 60.61 & 66.45 & 70.08 & 70.64 & 79.75 & \textbf{71.55} & 72.47 \\
& \textbf{Text-printed Image (Ours)}      & \textbf{90.43} & \textbf{62.24} & \textbf{68.59} & \textbf{86.24} & \textbf{77.61} & \textbf{93.38} & 67.09 & \textbf{77.94} \\
\cdashline{2-10}[4.0pt/2.5pt]
& \cellcolor{gray!15} GT-Image (Oracle) & \cellcolor{gray!15} 93.51 & \cellcolor{gray!15} 61.63 & \cellcolor{gray!15} 70.54 & \cellcolor{gray!15} 86.76 & \cellcolor{gray!15} 78.77 & \cellcolor{gray!15} 93.89 & \cellcolor{gray!15} 75.16 & \cellcolor{gray!15} 80.04 \\
\midrule
\midrule
\multirow{5}{*}{\rotatebox[origin=c]{90}{LLaMA Vision}}
& Pretrained                   & 50.77 & 25.99 & 58.32 & 22.28 & 46.98 & 80.83 & 34.33 & 45.64 \\
\cdashline{2-10}[4.0pt/2.5pt]
& Text-only                   & 66.91 & 48.28 & 62.19 & 46.68 & 60.51 & 83.38 & 41.56 & 58.50 \\
& Text-to-Image                   & 86.81 & 61.18 & 67.96 & 39.04 & 52.34 & 66.78 & 48.44 & 60.36 \\
& \textbf{Text-printed Image (Ours)}     & \textbf{90.93} & \textbf{62.65} & \textbf{70.44} & \textbf{73.28} & \textbf{65.36} & \textbf{90.84} & \textbf{52.37} & \textbf{72.27} \\
\cdashline{2-10}[4.0pt/2.5pt]
& \cellcolor{gray!15} GT-Image (Oracle) & \cellcolor{gray!15} 93.65 & \cellcolor{gray!15} 63.04 & \cellcolor{gray!15} 72.32 & \cellcolor{gray!15} 76.48 & \cellcolor{gray!15} 67.90 & \cellcolor{gray!15} 92.47 & \cellcolor{gray!15} 55.16 & \cellcolor{gray!15} 74.43 \\
\bottomrule
\end{tabular}
}
\vspace{-3mm}
\end{table*}

\section{Experiments}\label{sec:experiments}

We evaluate the effectiveness of TPI when synthetic textual descriptions are available from three perspectives:\\
(i) \emph{How closely can TPI training approach image-based training in downstream performance?} (Section~\ref{subsec:eval_benchmark}): We train models using textual descriptions derived from ground-truth images and compare them with models trained directly on the ground-truth images.\\
(ii) \emph{How does training with TPI change model behavior?} (Section~\ref{subsec:eval_behavior}): We analyze the behaviors of trained models in depth by comparing the output distributions and internal representations.\\
(iii) \emph{What factors enable effective learning with TPI?} (Section~\ref{subsec:eval_analysis}): We conduct various analyses to characterize when and why TPI training works.

\subsection{Experimental Setup}

\noindent
\textbf{Datasets.}
We use the following three types of datasets:
\begin{enumerate}
    \item \emph{General VQA}: We use three commonly used datasets: ScienceQA~\cite{marino2019ok}, OK-VQA~\cite{marino2019ok}, and VizWiz~\cite{gurari2018vizwiz}.
    \item \emph{Text VQA}: We also use three datasets that utilize images containing text: ChartQA~\cite{masry2022chartqa}, InfoVQA~\cite{mathew2022infographicvqa}, and DocVQA~\cite{mathew2021docvqa}. These tasks contain images like charts and posters that are difficult to represent with text alone.
    \item \emph{Domain-specific VQA}: To test whether TPI training transfers to unseen domains, we use DriveLM~\cite{sima2024drivelm}, a VQA dataset for autonomous driving. It contains vehicle-view images and corresponding questions.
\end{enumerate}
For General VQA and Text VQA, we evaluate performance using the lmms-eval library~\cite{zhang2024lmmsevalrealitycheckevaluation}. For DriveLM, we follow the official implementation and adopt evaluation by GPT-4.



\noindent
\textbf{TPI Generation Settings.}
For a fair comparison with image-based training, we first automatically generate textual descriptions from each ground-truth image using Qwen2.5-VL-32B~\cite{wang2024qwen2}.
We simply provide the model with an image and a question to generate an image description.
We note that the GT images are used solely to generate textual descriptions and are not required for practical use.
For each description, we render text using Python and the Pillow library~\cite{clark2015pillow}. 
Each TPI is an RGB image of size 336×336 px with a white background and black text. The font size is capped at 32 pt and is downscaled as needed to fit within the canvas (see Appendix~\ref{appendix:TPI_gen_setting} for details). 


\noindent
\textbf{Models and Training Setup.}
We use four LVLMs: LLaVA1.5 7B/13B~\cite{liu2023visual}, Qwen2.5 VL 7B Instruct~\cite{wang2024qwen2}, and Llama 3.2 11B Vision~\cite{dubey2024llama}. 
All models are fine-tuned with LoRA.
Detailed setups are in Appendix~\ref{appendix:training_details}.

\noindent
\textbf{Baselines.}
We use three baselines for comparison:
(i) \emph{Text-only training}, which directly uses the given textual description as training data.
(ii) \emph{Text-to-Image (T2I)}, which involves generating images from the textual descriptions using T2I model and then using those images for training. We use Stable Diffusion XL 1.0~\cite{podell2024sdxl} as T2I model. We sample with 25 denoising steps and a guidance scale of 5.0.
(iii) \emph{Ground-Truth Image (GT-Image, Oracle)}, which uses the originally paired image for each text for training, as a performance upper bound.

\subsection{Downstream Performance Improvement}\label{subsec:eval_benchmark}
We first evaluate training effectiveness using benchmark scores.
The results in Table~\ref{tab:vlm_benchmark} show that (i) the modality gap exists in text-only training, and (ii) TPI reduces this gap and improves performance.


\vspace{-12pt}
\paragraph{Modality gap.}
The results for LLaVA 7B/13B and LLaMA Vision show significant performance differences between \emph{Text-only} and \emph{GT-Image}, supporting that the improvements from text-only training are fundamentally constrained by the image–text modality gap.
Although the loss decreased during text-only training, trained models fail to handle image inputs effectively at test time.
In contrast, for Qwen VL, the performance difference between \emph{Text-only} and \emph{GT-Image} is not substantial. We attribute this to Qwen's higher intrinsic performance, which makes the gap less apparent and potentially a lower bias in its image encoder.


\vspace{-12pt}
\paragraph{Effectiveness of TPI.}
TPI outperforms baseline methods, indicating that it effectively mitigates the modality gap while preserving semantic fidelity.
Across all models, TPI clearly exceeds \emph{Text-only} and, on some tasks, approaches \emph{GT-Image}.
\emph{T2I} in particular underperforms on Text VQA, which is consistent with prior findings~\cite{li2024enhanced}.
We attribute this to the difficulty of diffusion models in generating images that faithfully reflect the given text.
We provide a detailed quantitative evaluation for faithfulness in Section~\ref{subsec:eval_analysis}.






\begin{figure*}[t]
  \centering
  \includegraphics[width=\linewidth]{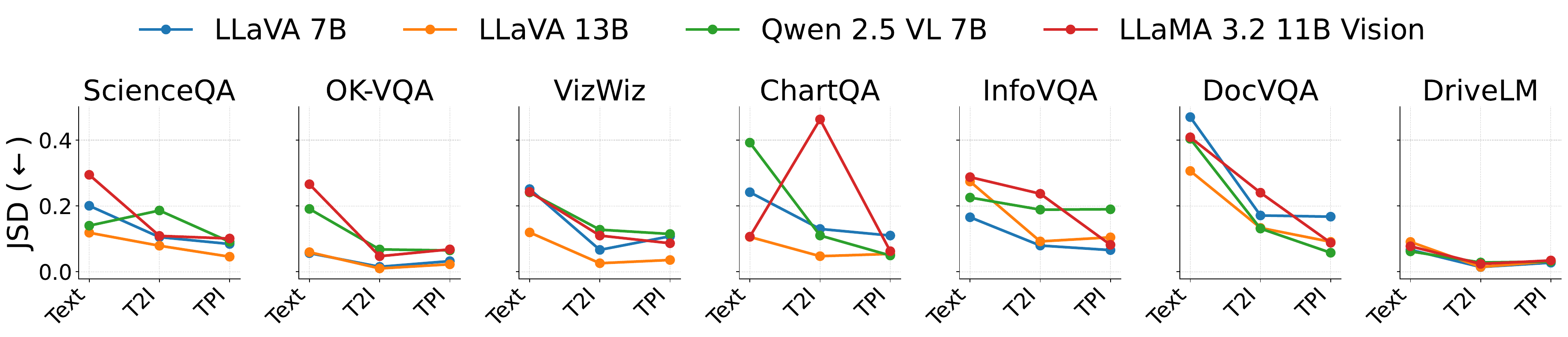}
  \caption{
  \textbf{TPI training yields output distributions most similar to models trained on GT-Images.}
  We compare the similarity to the model trained on \emph{GT-Image} in output distributions.
  Each value represents the JS divergence on the test split.
  }
  \label{fig:output_dist_analysis}
  \vspace{-6mm}
\end{figure*}

\begin{figure}[t]
  \centering
  \includegraphics[width=\linewidth]{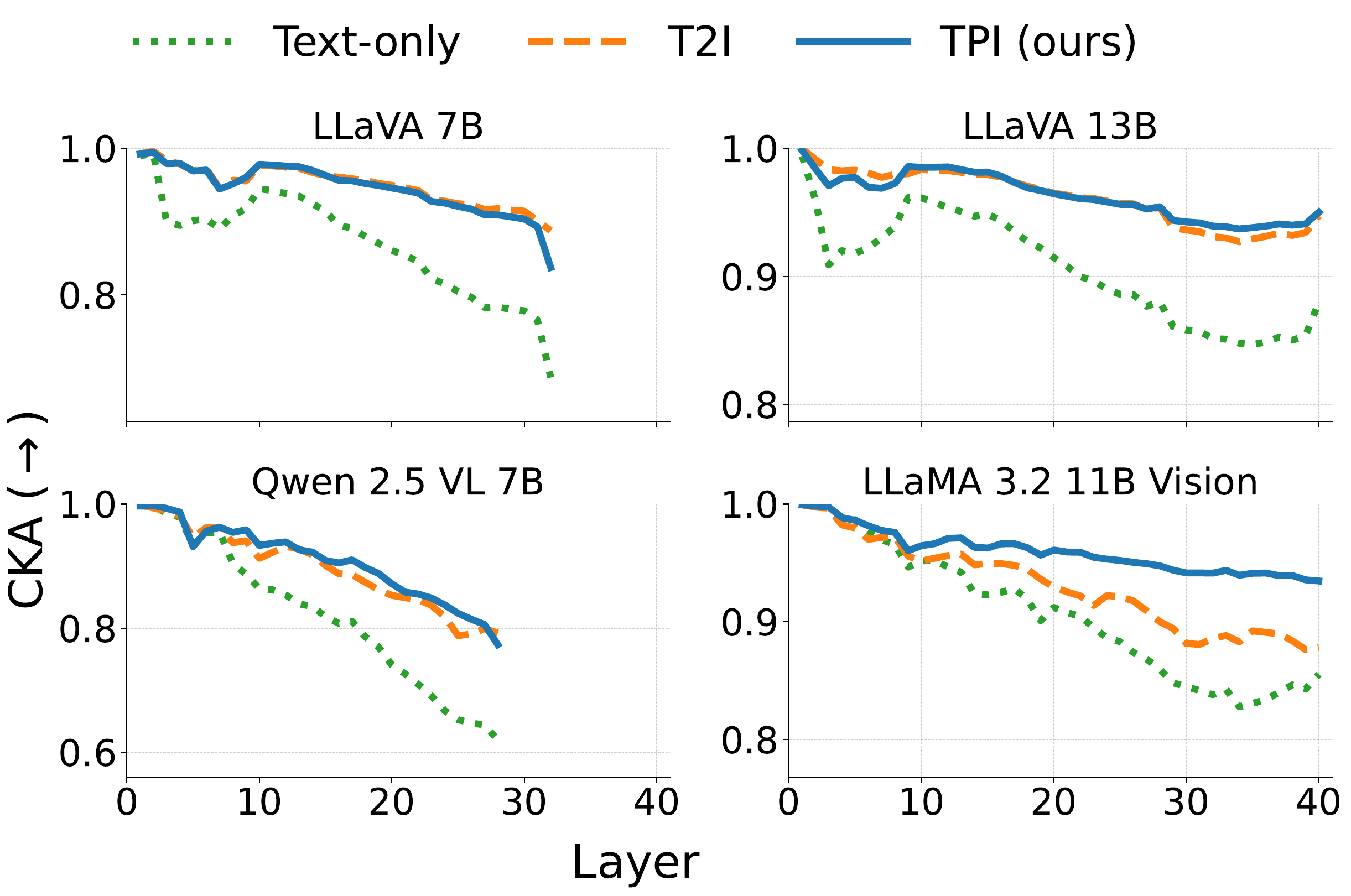}
  \caption{
  \textbf{Text-only training causes substantial drift in intermediate representations, while TPI mitigates it.}
  We compare the similarity to the model trained on \emph{GT-Image} in intermediate representations by computing the layer-wise CKA.
  }
  \label{fig:intermediate_feature_comparison}
  \vspace{-3mm}
\end{figure}



\subsection{Model Behavior Analyses}\label{subsec:eval_behavior}
To understand how TPI affects model behavior, we compare a model trained on \emph{GT-Image} with one trained on TPI.

\subsubsection{Output Distribution Comparison}

We assess similarity between output distributions using the JS divergence for each task’s test split to clarify behavioral differences that accuracy may obscure.
As shown in Figure~\ref{fig:output_dist_analysis}, TPI yields smaller divergences than the baselines, indicating effective training also at the level of output distributions. In contrast, \emph{Text-only} exhibits consistently higher JS divergence, suggesting that training without images alters model behavior, even when downstream task scores may appear reasonable.
\emph{T2I} shows notably lower similarity on Text VQA, which is consistent with the results in Table~\ref{tab:vlm_benchmark}.

\subsubsection{Intermediate Representation Comparison}

To examine how training inputs alter internal behavior, we compare intermediate representations. 
For each task’s test split, we extract the hidden state at the \emph{last token} position and compute \emph{Centered Kernel Alignment (CKA)}~\cite{kornblith2019similarity}, which quantifies similarity between representation matrices, layer by layer. We report the average value across all tasks.


As shown in Figure~\ref{fig:intermediate_feature_comparison}, \emph{Text-only} induces substantially larger shifts in intermediate representations compared with \emph{T2I} and TPI.
Notably, while performance improvements by \emph{Text-only} are comparable to \emph{GT-Image} for Qwen VL in Table~\ref{tab:vlm_benchmark}, the drift in intermediate representations is pronounced.
This suggests that training without images can alter behaviors and representations in ways that benchmark scores alone fails to capture.
In contrast, TPI mitigates this drift and better preserves the geometric structure of hidden states learned from the original visual modality.

\begin{figure}[t]
  \centering
  \includegraphics[width=\linewidth]{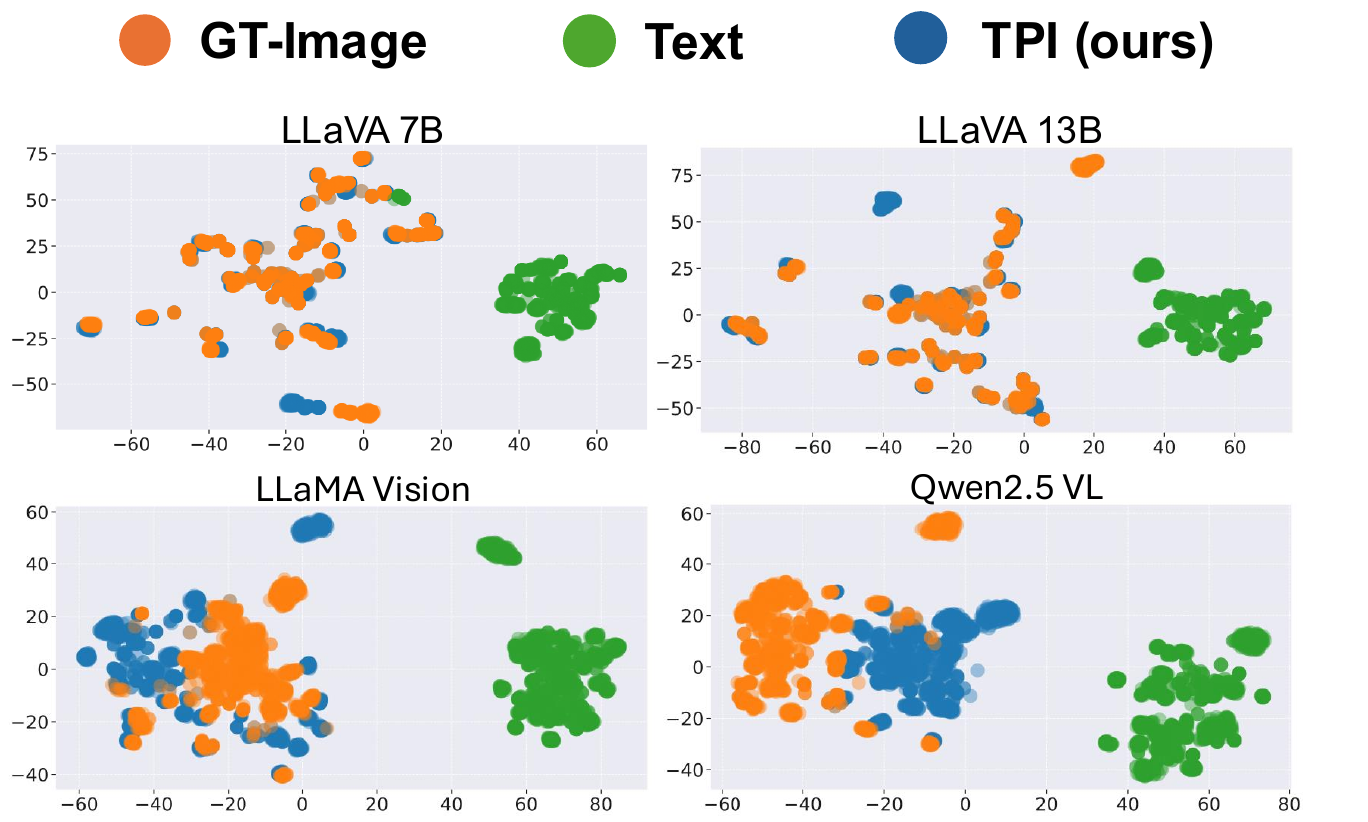}
  \caption{
  \textbf{TPI reduces the image–text modality gap.}
  We visualize t-SNE embeddings of intermediate features.
  While \emph{Text-only} produces a large separation from image features, TPI aligns more closely with the visual manifold.
  }
  \label{fig:tSNE}
  \vspace{-4mm}
\end{figure}

\subsection{Analysis of TPI Effectiveness}\label{subsec:eval_analysis}

\subsubsection{t-SNE Visualization of the Modality Gap}
To further investigate whether TPI effectively mitigates the image–text modality gap, we visualize the latent feature distributions. 
We feed the training examples into the pretrained model, extract the intermediate representations at the final token, and then apply t-SNE to these features.


Figure~\ref{fig:tSNE} shows the results in the ScienceQA task. The representations of \emph{GT-Image} and TPI are closely located in the same region of the feature space, while \emph{Text-only} form a clearly separated cluster.
These qualitative results support our claim that TPI effectively bridges the modality gap by projecting textual information into the image modality.

\begin{table}[t]
\centering
\caption{\textbf{Although T2I often fails to follow the provided prompts faithfully, TPI maintains high relevance.} We compute the Relevance Score, which measures how well each image aligns with the corresponding paired query and response.}
\label{tab:relevance_score}
\vspace{-1mm}
\setlength{\tabcolsep}{6pt}
\renewcommand{\arraystretch}{1.1}
\footnotesize
\resizebox{\linewidth}{!}{
\begin{tabular}{l:>{\columncolor{gray!12}}c:cc}
\toprule
Dataset & \cellcolor{gray!12}GT-Img. & T2I & \textbf{TPI (ours)} \\
\midrule
ScienceQA & 70.15 & 34.75 (49.5\%) & \textbf{57.32 (81.7\%)} \\
OK-VQA    & 69.27 & 54.38 (78.5\%) & \textbf{55.08 (79.5\%)} \\
VizWiz    & 65.79 & 44.49 (67.6\%) & \textbf{54.92 (83.5\%)} \\
ChartQA   & 79.22 & 20.78 (26.2\%) & \textbf{89.01 (112.4\%)} \\
InfoVQA   & 80.13 & 21.07 (26.3\%) & \textbf{80.53 (100.5\%)} \\
DocVQA    & 94.02 & 21.93 (23.3\%) & \textbf{79.54 (84.6\%)} \\
DriveLM   & 31.91 & \textbf{29.75 (93.2\%)} & 29.55 (92.6\%) \\
\midrule
Avg.      & 70.07 & 32.45 (46.3\%) & \textbf{63.61 (90.8\%)} \\
\bottomrule
\end{tabular}
}
\vspace{-2mm}
\end{table}

\subsubsection{Relevance Score Evaluation}\label{subsubsec:relevance_score}

To evaluate how TPI preserves given text semantics, we compute the relevance between synthetic images and their paired questions and answers. Following \textsc{VQAScore}~\cite{lin2024evaluating}, we define a \textsc{Relevance Score} as a probability that measures the consistency:
\begin{equation*}
\begin{split}
\Pr\Bigl(
&\text{``Yes''}\Bigm|\textit{image},\; \text{``Is the image relevant to the following} \\
&\text{Question and Answer? Please answer Yes or No.}
\Bigr).
\end{split}
\end{equation*}
A higher score indicates stronger alignment between the image and QA.
We compute the score using Qwen2.5-VL-32B~\cite{wang2024qwen2}, a sufficiently trained model. 


As shown in Table~\ref{tab:relevance_score}, the relevance scores of TPI are higher than those for T2I. 
This confirms our claim that TPI can preserve text semantics while T2I often struggles to remain faithful to the text, and the generated images can include elements that conflict with the QA.




\subsubsection{Impact of OCR Ability on TPI Effectiveness}
Since interpreting TPI requires recognizing text embedded in images, we examine how OCR capability relates to effectiveness.
We assess OCR ability using \textit{OCRBench}~\cite{liu2024ocrbench} and \textit{TextVQA}~\cite{singh2019towards}.
We quantify the effectiveness of TPI training by computing the Gap Ratio (GR): $\frac{\mathrm{TPI} - \mathrm{Pretrained}}{\mathrm{GT-Img.} - \mathrm{Pretrained}}$. We report the average score across all tasks.


Table~\ref{tab:ocr_gcr} shows that models with higher OCR scores tend to exhibit larger GRs, indicating OCR ability correlates with TPI effectiveness.
Although Qwen attains the highest OCR score, its strong \textit{Pretrained} baseline makes the denominator $(\mathrm{GT-Img.}-\mathrm{Pretrained})$ small, yielding a lower GR.
These observations suggest that models with at least moderate OCR ability benefit more from TPI. Nevertheless, TPI still outperforms \textit{Text-only} even for LLaVA-7B/13B, demonstrating that TPI can be effective even when OCR ability is limited.

\begin{table}[t]
\centering
\caption{
\textbf{Relationship between OCR capability and the effectiveness of TPI.}
Gap Ratio represents the fraction of \emph{GT-Img.} gains that TPI training recovers.
}
\label{tab:ocr_gcr}
\small
\setlength{\tabcolsep}{6pt}
\renewcommand{\arraystretch}{1.15}
\begin{tabular}{lcc>{\columncolor{gray!12}}c}
\toprule
 & OCRBench & TextVQA & \multicolumn{1}{c}{\textbf{Gap Ratio}} \\
\midrule
LLaVA 7B      & 20.3 & 47.9 & 64\% \\
LLaVA 13B     & 22.6 & 52.8 & 61\% \\
Qwen2.5 VL    & 83.7 & 82.9 & 79\% \\
Llama Vision  & 75.2 & 73.4 & 92\% \\
\bottomrule
\end{tabular}
\vspace{-2mm}
\end{table}

\begin{table}[t]
\centering
\caption{Image generation time for 6{,}218 descriptions.}
\label{tab:image_gen_time}
\resizebox{0.9\linewidth}{!}{
\begin{tabular}{lcc}
\toprule
Method & Total time [s] & Images per second \\
\midrule
T2I (1 H100 GPU) & 39347 & 0.16 \\
\rowcolor{gray!15} \textbf{TPI (CPU-only, ours)} & \textbf{40} & \textbf{154.40} \\
\bottomrule
\end{tabular}
}
\vspace{-3mm}
\end{table}

\subsubsection{Image generation efficiency}\label{subsubsec:efficiency}
To compare the computational cost of data construction, we measure the wall-clock time required to generate images.
For TPI, we use Intel Xeon Platinum 8480+ CPUs (32 cores, 64 threads), and for T2I, we use a single NVIDIA H100 GPU.
Table~\ref{tab:image_gen_time} reports the results on 6{,}218 ScienceQA images: T2I takes 10.9 hours, whereas TPI finishes in 40.27 seconds.
This result indicates that TPI enables substantially faster data construction without any GPU, making it a practical choice for large-scale text-centric training.

%% file: sec/5_augmentation.tex
\section{Text-Centric Data Augmentation at Scale}\label{sec:practcal_application}

We present an exploratory study toward scalable text-centric training. 
We augment textual description at scale by using an LLM and then train with TPI.
We examine two regimes: (i) a low-resource setting, where only a small seed subset is available, to test whether LLM-augmented descriptions with TPI improve over the pretrained model; and (ii) a full-data setting, where the entire training set is available, to assess whether the same augmentation still yields gains over a baseline trained on the full original dataset.

\subsection{Augmentation Methods Setup}
We adapt an established augmentation method for LLMs~\cite{wang2023self} to the data generation for LVLMs with minimal modifications. From a given initial data pool, we randomly select eight demonstrations, which consist of a query, response, and description. We then provide them to the LLM to generate one new sample. The new sample is added to the data pool, and the procedure is repeated. We first generate 10,000 samples, then compute ROUGE-L against the existing dataset and filter out highly similar items with scores $\ge 0.8$. We use \texttt{GPT-4o-mini} for generation, and the training setup is identical to Section~\ref{sec:experiments}. For details on data generation, see Appendix~\ref{appendix:details_data_augmentation}.

\begin{table}[t]
\centering
\caption{Performance of data augmentation with a small amount of original data via text-centric training.}
\label{tab:augmentation}
\setlength{\tabcolsep}{4pt}
\renewcommand{\arraystretch}{1.1}
\footnotesize
\resizebox{1.0\linewidth}{!}{
\begin{tabular}{>{\centering\arraybackslash}m{3mm} l *{4}{r}}
\toprule
& & \multicolumn{4}{c}{Models} \\
\cmidrule(lr){3-6}
& Methods & LLaVA 7B & LLaVA 13B & Qwen & LLaMA \\
\midrule

\multirow{8}{*}{\rotatebox{90}{ScienceQA}}
  & \cellcolor{gray!15} Pretrained   & \cellcolor{gray!15} 66.09 & \cellcolor{gray!15} 71.39 & \cellcolor{gray!15} 76.65 & \cellcolor{gray!15} 50.77 \\
\cdashline{2-6}[4.0pt/2.5pt]
& \multicolumn{5}{l}{Orig. (1\%)} \\
& + Text-only           & \textbf{65.44} & 70.15 & 86.96 & 66.04 \\
& + T2I            & 64.01 & \underline{71.05} & \underline{86.81} & \textbf{87.36} \\
& + \textbf{TPI (Ours)}   & \underline{64.20} & \textbf{70.40} & \textbf{88.25} & \underline{86.96} \\
\cmidrule(lr){2-6}
& \multicolumn{5}{l}{Orig. (10\%)} \\
& + Text-only          & 62.02 & 71.24 & 88.15 & 75.21 \\
& + T2I          & \textbf{69.36} & \underline{72.78} & \underline{89.94} & \underline{89.29} \\
& + \textbf{TPI (Ours)}  & \underline{68.57} & \textbf{73.28} & \textbf{90.33} & \textbf{89.39} \\
\midrule

\multirow{8}{*}{\rotatebox{90}{OK-VQA}}
  & \cellcolor{gray!15} Pretrained  & \cellcolor{gray!15} 49.87 & \cellcolor{gray!15} 52.49 & \cellcolor{gray!15} 43.88 & \cellcolor{gray!15} 25.99 \\
\cdashline{2-6}[4.0pt/2.5pt]
& \multicolumn{5}{l}{Orig. (1\%)} \\
& + Text-only          & \underline{55.03} & \underline{58.05} & 60.28 & 34.95 \\
& + T2I           & 52.00 & 55.87 & \underline{59.07} & \underline{54.09} \\
& + \textbf{TPI (Ours)}  & \textbf{55.62} & \textbf{61.04} & \textbf{61.94} & \textbf{57.23} \\
\cmidrule(lr){2-6}
& \multicolumn{5}{l}{Orig. (10\%)} \\
& + Text-only        & 57.85 & 62.03 & \textbf{63.19} & 42.46 \\
& + T2I          & \underline{58.78} & \underline{62.52} & 62.07 & \underline{59.66} \\
& + \textbf{TPI (Ours)} & \textbf{60.26} & \textbf{64.26} & \underline{62.81} & \textbf{60.18} \\
\bottomrule
\end{tabular}
}
\smallskip
\vspace{-3mm}
\end{table}

\subsection{Augmentation from Small Dataset}
To simulate data-scarce conditions, we use 1\% and 10\% subsets of the original dataset as the initial data pool. For training, we use a mixed dataset that combines the original subset with the synthesized data.
Results in Table~\ref{tab:augmentation} show that the performance of the trained model with the synthesized dataset exceeds that of the pretrained models. It suggests that augmentation via TPI is effective.
A particularly notable performance improvement was observed on OK-VQA. This is likely because OK-VQA is a task that depends on external knowledge, and data augmentation may have enabled the LLM to synthesize new data containing novel knowledge that is not included in the original subset.
Furthermore, in a comparison to baselines, TPI shows the greatest performance increase under most conditions, which is consistent with our earlier analyses in Section~\ref{sec:experiments} and supports the high quality of TPI as a training signal.

\subsection{Additional Augmentation from Full Dataset}
Next, we investigate whether TPI can achieve further performance improvements in a scenario where the full training dataset is available.
Results are shown in Table~\ref{tab:augmentation_full}. For ScienceQA, we observe up to about a 2-point increase, indicating that augmentation with TPI can still contribute even in near-saturation regimes.
In contrast, the improvement on OK-VQA is small.
This might be because the full dataset already contains a sufficiently diverse range of external knowledge, and the LLM was unable to generate enough supplementary synthetic data to complement it.

\vspace{-10pt}
\paragraph{Remark.} 
The significance of this experiment lies not only in demonstrating score improvements but in illustrating a promising direction for low-cost and practical data augmentation.
There is still significant room for development in data augmentation based on text-centric training. The augmentation method used in this experiment is a minimal adaptation of an existing method for LLMs. Thus, further improvements can be expected by developing methods specifically designed for text-centric training in LVLMs.

\begin{table}[t]
\centering
\caption{Performance of data augmentation with full original data via text-centric training.}
\label{tab:augmentation_full}
\setlength{\tabcolsep}{4pt}
\renewcommand{\arraystretch}{1.1}
\footnotesize
\resizebox{1.0\linewidth}{!}{
\begin{tabular}{>{\centering\arraybackslash}m{3mm} l *{4}{r}}
\toprule
& & \multicolumn{4}{c}{Models} \\
\cmidrule(lr){3-6}
& Methods & LLaVA 7B & LLaVA 13B & Qwen & LLaMA \\
\midrule

\multirow{4}{*}{\rotatebox{90}{ScienceQA}}
  & \cellcolor{gray!15} Orig. (100\%) & \cellcolor{gray!15} 78.78 & \cellcolor{gray!15} 80.12 & \cellcolor{gray!15} 93.51 & \cellcolor{gray!15} 93.65 \\
  \cdashline{2-6}[4.0pt/2.5pt]
  & + Text-only                    & 71.15 & 74.47 & 90.48 & 85.67 \\
  & + T2I                   & \underline{79.47} & \textbf{81.80} & \underline{94.99} & \underline{94.15} \\
  & + \textbf{TPI (Ours)}            & \textbf{80.27} & \underline{81.66} & \textbf{95.49} & \textbf{94.60} \\
\midrule

\multirow{4}{*}{\rotatebox{90}{OK-VQA}}
  & \cellcolor{gray!15} Orig. (100\%) & \cellcolor{gray!15} 62.12 & \cellcolor{gray!15} 64.53 & \cellcolor{gray!15} 61.63 & \cellcolor{gray!15} 63.04 \\
  \cdashline{2-6}[4.0pt/2.5pt]
  & + Text-only                    & 60.29 & 64.02 & \textbf{63.70} & 51.30 \\
  & + T2I                     & \underline{61.05} & \underline{64.55} & 62.13 & \textbf{63.48} \\
  & + \textbf{TPI (Ours)}            & \textbf{61.21} & \textbf{64.56} & \underline{62.55} & \underline{63.38} \\
\bottomrule
\end{tabular}
}
\smallskip
\end{table}

%% file: sec/6_discussion.tex
\section{Discussion and Limitation}\label{sec:limitation}
\paragraph{Generation setting of TPI.}
We conduct an extensive ablation on layout parameters, including font size, image resolution, and caption-generation method in Appendix~\ref{appendix:ablation}.
We did not observe substantial performance differences across these ablations. This suggests that, for learning, what matters is the semantics of the text rendered in the TPI rather than its typographic style or layout.

\vspace{-6pt}
\paragraph{Limitation.}
Our approach assumes that trained LVLMs possess certain visual understanding capabilities.
Because TPI provides properties such as shape and color only as text, a model that has not already acquired those visual concepts cannot reliably learn them from TPI alone. Also, TPI requires a minimum level of OCR ability.
In this sense, TPI implicitly assumes a setting where we start from a foundation model with broad visual knowledge and aim to specialize it toward task-specific capabilities, rather than teaching entirely new visual concepts from scratch.
However, given that such foundation models are becoming increasingly common, TPI is a promising low-cost and effective solution for text-centric training.

%% file: sec/7_conclusion.tex
\section{Conclusion}
We introduce TPI for effective text-centric training. TPI is lightweight and can be applied to a broad range of LVLM training pipelines without modification.
Our experiments show that TPI mitigates the image–text modality gap through multiple evaluations, including benchmark improvements, alignment of internal representations, and visualization analyses.
We further demonstrate that TPI offers better practical utility, such as preserving semantics, than T2I by comparing QA relevance and generation cost.
Finally, we apply TPI to a data augmentation scenario and demonstrate that it further improves performance.
Overall, our findings highlight TPI as a promising path toward scalable training for LVLMs.




%% file: sec/X_suppl.tex
\clearpage
\setcounter{page}{1}

\onecolumn
\maketitlesupplementary

\appendix

\section{Experimental Details}

\subsection{Compute Resources}
All experiments were conducted on an HPC cluster.
For most experiments, we use a single NVIDIA H100 GPU.
For generating synthetic images with T2I, we use a node equipped with 8× NVIDIA H100 GPUs.
The system uses NVIDIA driver 535.183.01 and CUDA 12.2.

\subsection{Training Details}\label{appendix:training_details}
\paragraph{Training data.}
For all tasks, we train models using the provided training split of each dataset. Input queries are formatted with the chat templates associated with each LVLM. To ensure fair comparison under \texttt{lmms-eval}~\cite{zhang2024lmmsevalrealitycheckevaluation}, we format the model responses to match the output format expected by the \texttt{lmms-eval} evaluation pipeline. For DriveLM~\cite{sima2024drivelm}, although the original setting uses six images per example, we simplify training by using only the single image corresponding to each QA pair.

\paragraph{Hyperparameters.}
We use the same training configuration for all models. We optimize with AdamW, set the connector learning rate to $1\times10^{-5}$, and apply a cosine learning-rate schedule with a warm-up ratio of 0.03, no weight decay, a global batch size of 4, and gradient accumulation of 4. We train for 3 epochs in \texttt{bfloat16} precision. All models are fine-tuned with LoRA with rank $r=256$ and scaling factor $\alpha=512$. Following prior work~\cite{zhou2024empirical}, we update only the parameters of the LLM, while keeping the visual encoder frozen.

\subsection{Details of Generating Textual Descriptions}\label{appendix:description_generation}
We automatically generate textual descriptions from ground-truth images using an LVLM. Specifically, we employ Qwen2.5-VL-32B~\cite{wang2024qwen2}, a well-trained LVLM. For each sample, we provide the ground-truth image and its corresponding QA pair as input to the model and instruct it to produce a textual description that is relevant to the given QA. The exact prompt used for this generation is shown below:

\begin{tcolorbox}[
    colback=gray!10,
    colframe=gray!40,
    boxrule=0.3pt,
    arc=1pt,
    left=4pt,
    right=4pt,
    top=4pt,
    bottom=4pt,
    width=\columnwidth,
    enhanced,
    breakable
]
\small\ttfamily
System: You are a highly skilled visual description assistant. Given an image and, optionally, a question and its answer, write a single-paragraph, highly detailed, objective description of the image. Your goal is to capture all relevant visual elements in a way that would allow a reader to mentally reconstruct the image and, if applicable, answer the given question using only your description. Your description should include the type of scene (e.g., natural, diagram, poster, chart), the spatial layout of elements, visual attributes such as color, shape, and texture, any visible text or labels, and, if present, numerical or symbolic information. Do not include any interpretation, emotion, speculation, or bullet points. Keep the tone factual, precise, and comprehensive. Aim for approximately 100 words. \\

User:  \\
Question: \{question\} \\
Answer: \{answer\}
\end{tcolorbox}

\subsection{Relevance Score}
To quantify how faithfully a synthetic image matches its paired textual supervision, we compute \emph{Relevance Scores} by using Qwen2.5-VL-32B-Instruct.
Inference is performed with temperature = 0 and max\_tokens = 1 to obtain deterministic Yes or No predictions.
Images are attached through the official Qwen chat template. Specifically, we also use the following prompt: 
\begin{tcolorbox}[
    colback=gray!10,
    colframe=gray!40,
    boxrule=0.3pt,
    arc=1pt,
    left=4pt,
    right=4pt,
    top=4pt,
    bottom=4pt,
    width=\columnwidth,
    enhanced,
    breakable
]
\small\ttfamily
System: Output only Yes or No.\\
User: Is the image relevant to the following Q\&A?\\
Question: \{q\}\\
Answer: \{a\}
\end{tcolorbox}

For each prompt, we compute the probability of generating ``yes'' and ``no'' from the first tokens.

\subsection{t-SNE visualization}
For t-SNE visualizations, we first pass inputs constructed by each method (GT-Image, TPI, and Text-only) through the model and extract the hidden states from all transformer layers. 
For each input, we take the hidden vector at the last token position as the layer-wise representation and apply t-SNE to obtain a 2D embedding.
For hyperparameters, we set perplexity = 100.0, learning rate = 1000.0, and number of iterations = 5000.
Specifically, Figure~\ref{fig:tSNE} reports the t-SNE embeddings from layer 11 for LLaVA-7B, layer 10 for LLaVA-13B, and layer 20 for both Qwen2.5-VL and LLaMA~3.2~Vision.

\section{Details of TPI Generation Settings}\label{appendix:TPI_gen_setting}
In this section, we describe the details of generating synthetic images for TPI.
For a fair comparison with image-based training, we first automatically generate textual descriptions from each ground-truth image as described in Section~\ref{appendix:description_generation}.
We note that only these textual descriptions are used for training; original images are not required in practical use.
For each description, we render text using Python and the Pillow library~\cite{clark2015pillow}. 

\paragraph{Layout parameters.}
Each TPI is an RGB image of size 336×336 pixels.
By default, we use a plain white background and black text, and render with a TrueType font (DejaVu Sans).
For a given text, we perform a top–down search over font sizes: starting from a default font size (32 pt) and
decreasing it in steps until the wrapped text fits within the target canvas. At
each candidate size, we construct lines by greedy word-wrapping so that the
width of each line does not exceed the available width
(\texttt{img\_width} minus horizontal padding). We then estimate the line height
and total height of the wrapped text, including a fixed line spacing. If both
the maximum line width and total text height fit within the canvas after
respecting padding on all sides, we accept this font size and render the text.

\section{Details of Data Augmentation}\label{appendix:details_data_augmentation}

In this section, we describe the data augmentation setup used in Section~\ref{sec:practcal_application}.

\paragraph{Pipeline overview.}
We follow the Self-Instruct framework~\cite{wang2023self} to generate additional training data for our text-centric setting.
First, for each seed example, we generate an image caption from the associated image and use these captioned examples as the initial pool.
Then, at each iteration, we randomly sample $8$ examples from the pool as demonstrations and ask the LLM to generate \emph{one} new example.
We check whether the generated example overlaps with the existing pool using ROUGE-L.
If the example is not considered a duplicate, we add it to the pool and repeat the same procedure to generate the next example.
We run this generation loop for $10{,}000$ iterations for each task.

\begin{table}[t]
\centering
\small
\begin{tabular}{lcc}
\toprule
\textbf{Setting} & \textbf{ScienceQA} & \textbf{OK-VQA} \\
\midrule
Orig.\ (1\%)   & 6{,}965 & 6{,}194 \\
Orig.\ (10\%)  & 6{,}513 & 3{,}927 \\
Orig.\ (100\%) & 7{,}118 & 1{,}317 \\
\bottomrule
\end{tabular}
\caption{Number of generated augmented examples for each dataset and training fraction.}
\label{tab:augmented_datasets}
\end{table}

Table~\ref{tab:augmented_datasets} summarizes the number of generated augmented examples.
In particular, the number of generated examples for OK-VQA in the 100\% setting is relatively small.
We hypothesize that this is because the original OK-VQA dataset is already quite comprehensive, so there is limited room for the LLM to propose new, non-duplicate knowledge.
This also explains why the performance gains from augmentation are modest in this setting.

\paragraph{Model and prompt.}
For data augmentation, we use \texttt{gpt-4o-mini} as the generation model.
We set the temperature to $0.7$ to encourage diversity in the generated examples.
To strictly control the output format, we instruct the model to return a single JSON object that matches a predefined schema.
The core prompt is as follows:

\begin{tcolorbox}[
colback=gray!10,
colframe=gray!40,
boxrule=0.3pt,
arc=1pt,
left=4pt,
right=4pt,
top=4pt,
bottom=4pt,
width=\columnwidth,
enhanced,
breakable
]
\small\ttfamily
System: Return exactly one JSON object that validates against the given schema. No extra text. \\
User: Here are seed examples (one JSON per line): \\

\{demo\} \\

Produce ONE new and diverse example that is not copied. Output only the JSON object.
\end{tcolorbox}

There is still substantial room to tune this text-centric augmentation process.
For example, one could add extra instructions that modify only specific parts of the image description (such as changing certain objects) to obtain more diverse captions and questions.

\paragraph{Duplicate detection.}
We use ROUGE-L to detect duplicate-like examples.
For each newly generated example, we first convert it into a single canonical text string.
We then compute the ROUGE-L F1 score between this string and the canonical text of every example in the pool.
If the best ROUGE-L F1 score is greater than or equal to a threshold of $0.70$, we treat the example as duplicate-like and discard it.
This heuristic is conservative: even if many words overlap, changing a small number of key words (for example, replacing “cat’’ with “dog’’) can change the underlying question.
Such semantic changes are not always fully captured by ROUGE-L alone, so more advanced duplicate detection methods could further improve this step.

\paragraph{Training setup.}
For SFT, we use exactly the same training configuration as in the main experiments.
In all cases, we perform SFT with LoRA for $3$ epochs on the (original + augmented) training data.

\section{Ablation Study}\label{appendix:ablation}
In this section, we conduct an ablation study on how TPIs are generated.

\subsection{Font Size}

\begin{table*}[t]
\centering
\caption{Comparison of font sizes. Each value shows accuracy (\%) for different font sizes in TPI.}
\label{tab:ablate_fontsize}
\setlength{\tabcolsep}{4pt}
\renewcommand{\arraystretch}{1.1}
\begin{tabular*}{0.80\textwidth}{@{\extracolsep{\fill}}lcccc}
\toprule
& \multicolumn{4}{c}{\textbf{Font Size}} \\
Model & 4 & 16 & 32 & 64 \\
\midrule
\multicolumn{5}{l}{\emph{ScienceQA}}\\
\midrule
LLaVA 7B      & 73.57 & \textbf{74.27} & 73.87 & 73.76 \\
LLaVA 13B     & 75.20 & 76.00         & \textbf{76.60} & 75.95 \\
Qwen2.5 VL    & 89.69 & \textbf{91.22} & 90.68 & 88.00 \\
LLaMA Vision  & 86.25 & \textbf{90.33} & 89.99 & 87.80 \\
\hdashline
Avg.          & 81.18 & \textbf{82.95} & 82.78 & 81.38 \\
\midrule
\midrule
\multicolumn{5}{l}{\emph{OK-VQA}}\\
\midrule
LLaVA 7B      & 59.98 & 60.52 & \textbf{60.54} & 60.11 \\
LLaVA 13B     & 63.96 & \textbf{64.81} & 64.53 & 63.86 \\
Qwen2.5 VL    & 62.32 & 62.56 & \textbf{63.08} & 62.47 \\
LLaMA Vision  & 61.54 & \textbf{62.47} & 62.27 & 62.37 \\
\hdashline
Avg.          & 61.95 & 62.59 & \textbf{62.60} & 62.20 \\
\midrule
\midrule
\multicolumn{5}{l}{\emph{VizWiz}}\\
\midrule
LLaVA 7B      & 56.20 & \textbf{62.59} & 60.92 & 57.16 \\
LLaVA 13B     & 57.87 & \textbf{63.63} & 62.41 & 57.16 \\
Qwen2.5 VL    & 64.46 & \textbf{68.50} & 68.11 & 65.63 \\
LLaMA Vision  & 63.39 & \textbf{70.41} & 68.85 & 63.76 \\
\hdashline
Avg.          & 60.48 & \textbf{66.28} & 65.07 & 60.93 \\
\bottomrule
\end{tabular*}
\end{table*}

We evaluate how the font size of the text in TPI affects training. We consider four fixed font sizes: 4, 16, 32, and 64. In the main experiments in Section~\ref{sec:experiments}, we instead use a default font size of 32 and reduce it only when the text does not fit into the image, so these fixed sizes are not used there. All other training settings are exactly the same as in Section~\ref{sec:experiments}.

\paragraph{Results.}
The results are shown in Table~\ref{tab:ablate_fontsize}. We observe that models train well with font sizes 16 and 32, while performance drops for sizes 4 and 64, especially on VizWiz. When the font size is too small, the model may fail to read the characters, particularly when the input text is long and dense. When the font size is too large, the text can overflow outside the image, causing part of the information to be lost and making the TPI image less informative. Based on these results, we recommend using moderate font sizes such as 16 or 32.

\subsection{Font Color}
\begin{table*}[t]
\centering
\caption{Comparison of font colors. Each value shows accuracy (\%) for different font colors in TPI. We did not observe any substantial performance differences across font colors.}
\label{tab:ablate_color}
\setlength{\tabcolsep}{4pt}
\renewcommand{\arraystretch}{1.1}
\begin{tabular*}{0.80\textwidth}{@{\extracolsep{\fill}}lcccccc}
\toprule
& \multicolumn{6}{c}{\textbf{Font Color}} \\
Model & Black & Blue & Green & Orange & Red & Yellow \\
\midrule
\multicolumn{7}{l}{\emph{ScienceQA}}\\
\midrule
LLaVA 7B      & 75.11 & 74.81 & 74.91 & 74.37 & \textbf{75.16} & 75.01 \\
LLaVA 13B     & 76.30 & 76.65 & 76.55 & \textbf{76.95} & 76.85 & 76.65 \\
Qwen2.5 VL    & 90.43 & 91.47 & 91.77 & \textbf{91.92} & 91.18 & 91.32 \\
LLaMA Vision  & \textbf{90.93} & 90.43 & 90.28 & 90.38 & 90.33 & \textbf{90.93} \\
\hdashline
Avg.          & 83.19 & 83.34 & 83.38 & 83.41 & 83.38 & \textbf{83.48} \\
\midrule
\midrule
\multicolumn{7}{l}{\emph{OK-VQA}}\\
\midrule
LLaVA 7B      & \textbf{61.70} & 60.40 & 60.64 & 60.48 & 60.54 & 60.69 \\
LLaVA 13B     & 64.73 & \textbf{64.95} & 64.79 & 64.92 & 64.86 & 64.61 \\
Qwen2.5 VL    & 62.24 & \textbf{62.59} & 62.52 & 62.43 & 62.56 & 62.39 \\
LLaMA Vision  & 62.65 & \textbf{62.85} & 62.40 & 62.59 & 62.62 & 62.70 \\
\hdashline
Avg.          & \textbf{62.83} & 62.70 & 62.59 & 62.60 & 62.65 & 62.60 \\
\midrule
\midrule
\multicolumn{7}{l}{\emph{VizWiz}}\\
\midrule
LLaVA 7B      & 61.96 & 61.79 & 62.13 & 61.99 & 61.97 & \textbf{62.23} \\
LLaVA 13B     & \textbf{64.46} & 63.27 & 63.57 & 63.63 & 63.91 & 64.05 \\
Qwen2.5 VL    & 68.59 & 68.75 & \textbf{68.89} & 68.52 & 68.50 & 68.42 \\
LLaMA Vision  & \textbf{70.44} & 69.88 & 69.76 & 70.11 & 70.27 & 69.99 \\
\hdashline
Avg.          & \textbf{66.36} & 65.92 & 66.09 & 66.06 & 66.16 & 66.17 \\
\bottomrule
\end{tabular*}
\end{table*}

We evaluate how the font color of the text in TPI affects model training. Font color may influence the model’s ability to recognize characters. If some colors are easier for the model to read, using them could improve learning performance.

\paragraph{Setup.}
We consider six candidate font colors: \emph{black}, \emph{blue}, \emph{green}, \emph{orange}, \emph{red}, and \emph{yellow}. In the main experiments in Section~\ref{sec:experiments}, we use black as the default color. Apart from the font color, all training settings are exactly the same as in Section~\ref{sec:experiments}.

\paragraph{Results.}
The results are shown in Table~\ref{tab:ablate_color}. We do not observe large differences in performance across font colors. If anything, black yields the highest performance on average. A possible reason is that black text is also the most common in standard OCR datasets, so the model may find it easier to recognize. Based on these results, we recommend using black fonts when generating TPI. However, other colors are also acceptable, so in practice it may be preferable to choose a font color that matches the background while preserving sufficient contrast for readability.

\subsection{Generation Prompt}\label{subsec:ablate_prompt}
The amount of information contained in the given textual descriptions may affect the effectiveness of training. Therefore, we evaluate how the learning results change when we vary the prompts used to generate these descriptions. Specifically, we consider the following three types of prompts.

\paragraph{50 words.} This prompt restricts the description to at most 50 words. Although the description contains less information, it may retain only the most important content. The exact prompt we use is shown below:

\begin{tcolorbox}[
    colback=gray!10,
    colframe=gray!40,
    boxrule=0.3pt,
    arc=1pt,
    left=4pt,
    right=4pt,
    top=4pt,
    bottom=4pt,
    width=\columnwidth,
    enhanced,
    breakable,
    title={Prompt for 50 words.}
]
\small\ttfamily
You are an objective image captioning assistant. Describe strictly what you see in the image. Do NOT include apologies, judgments, context, or filler phrases. Use up to 50 words in your description. \\
\end{tcolorbox}

\paragraph{200 words.} This prompt restricts the description to at most 200 words, allowing a longer and more detailed explanation than the 50-word setting. The exact prompt we use is shown below:

\begin{tcolorbox}[
    colback=gray!10,
    colframe=gray!40,
    boxrule=0.3pt,
    arc=1pt,
    left=4pt,
    right=4pt,
    top=4pt,
    bottom=4pt,
    width=\columnwidth,
    enhanced,
    breakable,
    title={Prompt for 200 words.}
]
\small\ttfamily
You are an objective image captioning assistant. Describe strictly what you see in the image. Do NOT include apologies, judgments, context, or filler phrases. Use up to 200 words in your description, providing detailed coverage of objects, setting, colors, and actions. \\
\end{tcolorbox}

\paragraph{Rich.} This prompt instructs the model to generate as rich a description as possible. We expect the resulting descriptions to include many details and a large amount of information. The exact prompt we use is shown below:

\begin{tcolorbox}[
    colback=gray!10,
    colframe=gray!40,
    boxrule=0.3pt,
    arc=1pt,
    left=4pt,
    right=4pt,
    top=4pt,
    bottom=4pt,
    width=\columnwidth,
    enhanced,
    breakable,
    title={Prompt for Rich.}
]
\small\ttfamily
You are a creative image captioning assistant. Provide a rich, detailed description of the image, highlighting objects, setting, colors, textures, actions, emotions, and context. Use complete sentences and vivid language without apologies or filler phrases. \\
\end{tcolorbox}

\paragraph{24 words with QA.}
This prompt instructs the model to produce a QA-related description in no more than 24 words. By limiting the amount of information, the description is encouraged to focus more directly on objects that are relevant to answering the question. The exact prompt we use is shown below:

\begin{tcolorbox}[
    colback=gray!10,
    colframe=gray!40,
    boxrule=0.3pt,
    arc=1pt,
    left=4pt,
    right=4pt,
    top=4pt,
    bottom=4pt,
    width=\columnwidth,
    enhanced,
    breakable,
    title={Prompt for 24 words with QA.}
]
\small\ttfamily
You are an objective image-captioning assistant. Describe strictly what you see in the image in no more than 24 words. Ensure the caption contains the details required to answer the question. Write as one continuous paragraph-do NOT use bullet points, lists, apologies, opinions, or speculative content. \\
\end{tcolorbox}

\begin{table*}[t]
\centering
\caption{Comparison of generation prompts. Each value shows accuracy (\%) for different prompts in TPI.}
\label{tab:ablate_prompts}
\setlength{\tabcolsep}{4pt}
\renewcommand{\arraystretch}{1.1}
\begin{tabular*}{0.80\textwidth}{@{\extracolsep{\fill}}lccccc}
\toprule
& \multicolumn{5}{c}{\textbf{Description Length}} \\
Model & Default & 50 words & 200 words & Rich & 24 words with QA \\
\midrule
\multicolumn{6}{l}{\emph{ScienceQA}}\\
\midrule
LLaVA 7B      & 75.11 & 74.62 & 74.37 & \textbf{75.71} & 74.57 \\
LLaVA 13B     & 76.30 & 76.65 & 76.30 & \textbf{77.14} & 76.60 \\
Qwen2.5 VL    & 90.43 & 90.68 & \textbf{91.27} & 90.33 & 90.78 \\
LLaMA Vision  & \textbf{90.93} & 89.84 & 88.94 & 88.45 & 89.69 \\
\hdashline
Avg.          & \textbf{83.19} & 82.95 & 82.72 & 82.91 & 82.91 \\
\midrule
\midrule
\multicolumn{6}{l}{\emph{OK-VQA}}\\
\midrule
LLaVA 7B      & \textbf{61.70} & 60.24 & 60.48 & 59.91 & 59.84 \\
LLaVA 13B     & 64.73 & 64.72 & \textbf{64.83} & 64.58 & 64.52 \\
Qwen2.5 VL    & 62.24 & 62.05 & 62.23 & \textbf{62.62} & 62.37 \\
LLaMA Vision  & 62.65 & \textbf{62.77} & 62.63 & 62.51 & 62.53 \\
\hdashline
Avg.          & \textbf{62.83} & 62.44 & 62.54 & 62.40 & 62.32 \\
\midrule
\midrule
\multicolumn{6}{l}{\emph{VizWiz}}\\
\midrule
LLaVA 7B      & 61.96 & \textbf{62.06} & 60.75 & 61.86 & 61.90 \\
LLaVA 13B     & \textbf{64.46} & 62.93 & 63.26 & 62.96 & 63.90 \\
Qwen2.5 VL    & 68.59 & \textbf{69.82} & 68.48 & 68.96 & 68.51 \\
LLaMA Vision  & \textbf{70.44} & 69.22 & 69.49 & 68.62 & 69.02 \\
\hdashline
Avg.          & \textbf{66.36} & 66.01 & 65.50 & 65.60 & 65.83 \\
\bottomrule
\end{tabular*}
\end{table*}

\paragraph{Results.}
The results are shown in Table~\ref{tab:ablate_prompts}. \emph{Default} denotes the prompt used in our main experiments (see Section~\ref{appendix:description_generation} for details). We do not observe any large performance differences across the generation prompts. If anything, longer descriptions tend to yield slightly better performance. For example, 200words, rich, and default prompts achieve marginally higher scores than the others.

These results suggest that the amount of information in the textual description is not crucial for learning, as long as it contains the minimum set of correct information. In practice, there is little need to make the textual descriptions overly high-quality. Instead, increasing the diversity of QA pairs may be more beneficial for training.

\subsection{Generation Model}

\begin{table*}[ht]
\centering
\caption{Comparison of generation models. Each column corresponds to the Qwen model used for textual description generation.}
\label{tab:ablate_generator}
\setlength{\tabcolsep}{4pt}
\renewcommand{\arraystretch}{1.1}
\begin{tabular*}{0.80\textwidth}{@{\extracolsep{\fill}}lccc}
\toprule
& \multicolumn{3}{c}{\textbf{Generation Model}} \\
Model & Qwen2.5 VL 3B & Qwen2.5 VL 7B & Qwen2.5 VL 32B \\
\midrule
\multicolumn{4}{l}{\emph{ScienceQA}}\\
\midrule
LLaVA 7B      & \textbf{75.11} & 74.96 & \textbf{75.11} \\
LLaVA 13B     & 76.00 & \textbf{76.80} & 76.30 \\
Qwen2.5 VL    & 90.38 & \textbf{91.72} & 90.43 \\
LLaMA Vision  & 88.10 & 88.55 & \textbf{90.93} \\
\hdashline
Avg.          & 82.40 & 83.01 & \textbf{83.19} \\
\midrule
\midrule
\multicolumn{4}{l}{\emph{OK-VQA}}\\
\midrule
LLaVA 7B      & 59.61 & 60.31 & \textbf{61.70} \\
LLaVA 13B     & 64.24 & 64.67 & \textbf{64.73} \\
Qwen2.5 VL    & \textbf{62.57} & 62.49 & 62.24 \\
LLaMA Vision  & 60.48 & 61.02 & \textbf{62.65} \\
\hdashline
Avg.          & 61.72 & 62.12 & \textbf{62.83} \\
\midrule
\midrule
\multicolumn{4}{l}{\emph{VizWiz}}\\
\midrule
LLaVA 7B      & 60.27 & 58.82 & \textbf{61.96} \\
LLaVA 13B     & 61.57 & 60.75 & \textbf{64.46} \\
Qwen2.5 VL    & 69.09 & \textbf{71.05} & 68.59 \\
LLaMA Vision  & 69.54 & 69.35 & \textbf{70.44} \\
\hdashline
Avg.          & 65.12 & 64.99 & \textbf{66.36} \\
\bottomrule
\end{tabular*}
\end{table*}

In text-centric training, one of the main ways to build data is to automatically generate textual descriptions with LVLMs. Our experiments also follow this approach. In Section \ref{sec:experiments}, we automatically generate textual descriptions from ground-truth images using an LVLM. In Section \ref{sec:practcal_application}, we use the same strategy for generating new samples in the data-augmentation experiments.

In this section, we evaluate how differences in the generation model affect the final learning performance. To reduce the influence of model-specific writing bias and focus on the model’s core ability, we compare three models from the same Qwen family: Qwen2.5-VL-3B-Instruct, Qwen2.5-VL-7B-Instruct, and Qwen2.5-VL-32B-Instruct. This setup keeps the style of the generated text consistent while changing only the model size and capability, allowing us to isolate how these factors contribute to text-centric training.

The results are shown in Table~\ref{tab:ablate_generator}. We observe a clear trend: using a stronger model to generate textual descriptions leads to better VLM performance after training. In particular, textual descriptions produced by the strongest model, Qwen2.5-VL-32B-Instruct, achieve the highest average scores. A likely reason is that larger models can extract visual information more accurately and produce fewer mistakes in their descriptions.

Overall, these results suggest that textual descriptions do not need to be overly detailed. What matters more is how accurately they capture the information in the image. This contrasts with the ablation of generation prompts in Section \ref{subsec:ablate_prompt}, where changing the amount of information had little effect. However, changing the generation model does produce differences. This implies that correctness and relevance of the description are more important than its length. The finding also aligns with our observation that training with T2I-generated images, whose relevance scores are low, provides only limited improvements.

\section{Qualitative Examples}
To support our quantitative analysis, we present examples of the synthetic images used for training. In particular, to compare T2I and TPI more effectively, we highlight cases where T2I images deviate from the provided textual descriptions and consequently receive low relevance scores. 

\begin{figure*}[t]
  \centering
  \includegraphics[width=\linewidth]{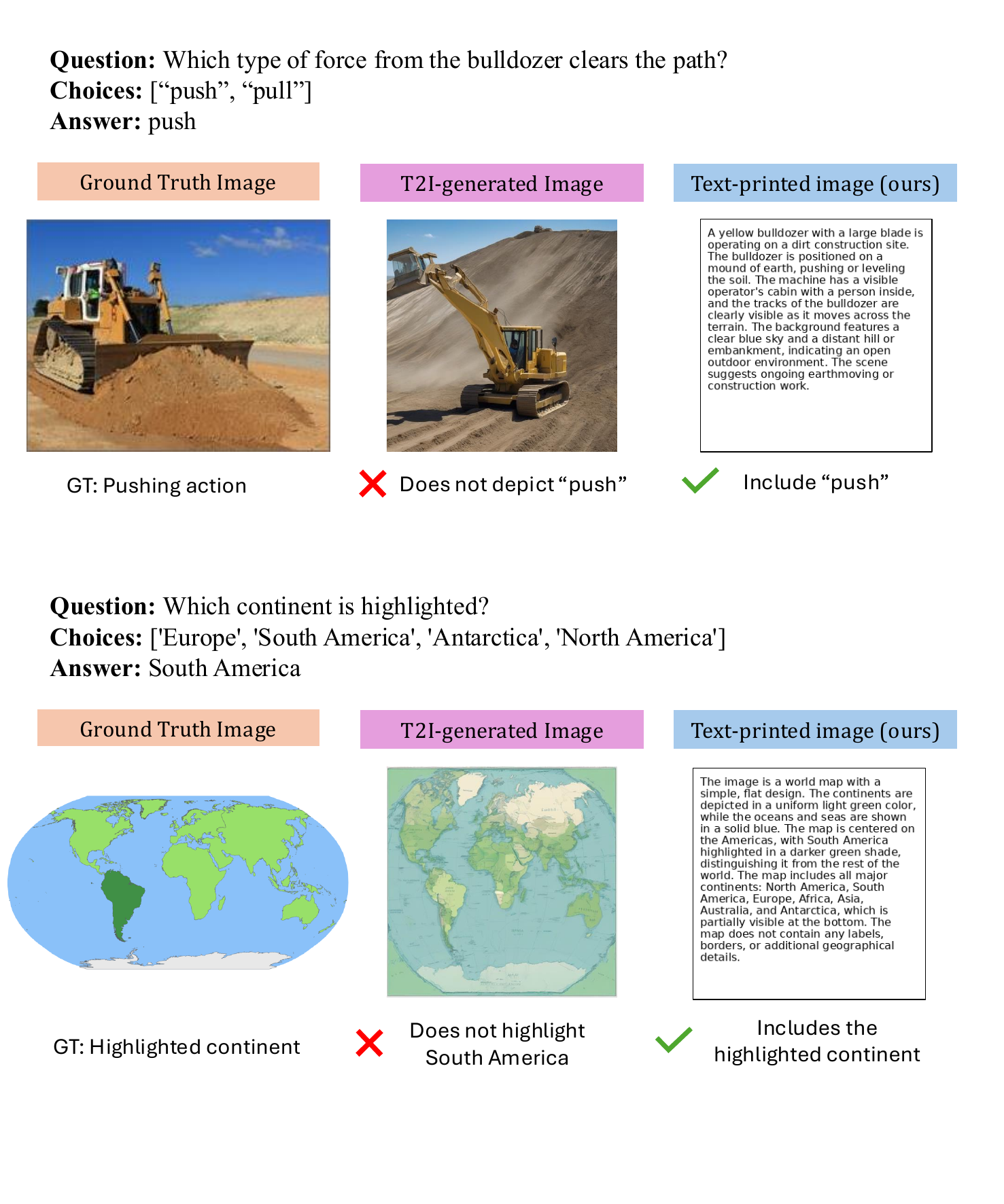}
  \caption{
  Qualitative comparison of synthetic images on ScienceQA.
  }
  \label{fig:qualitative_example_scienceQA}
\end{figure*}

\begin{figure*}[t]
  \centering
  \includegraphics[width=\linewidth]{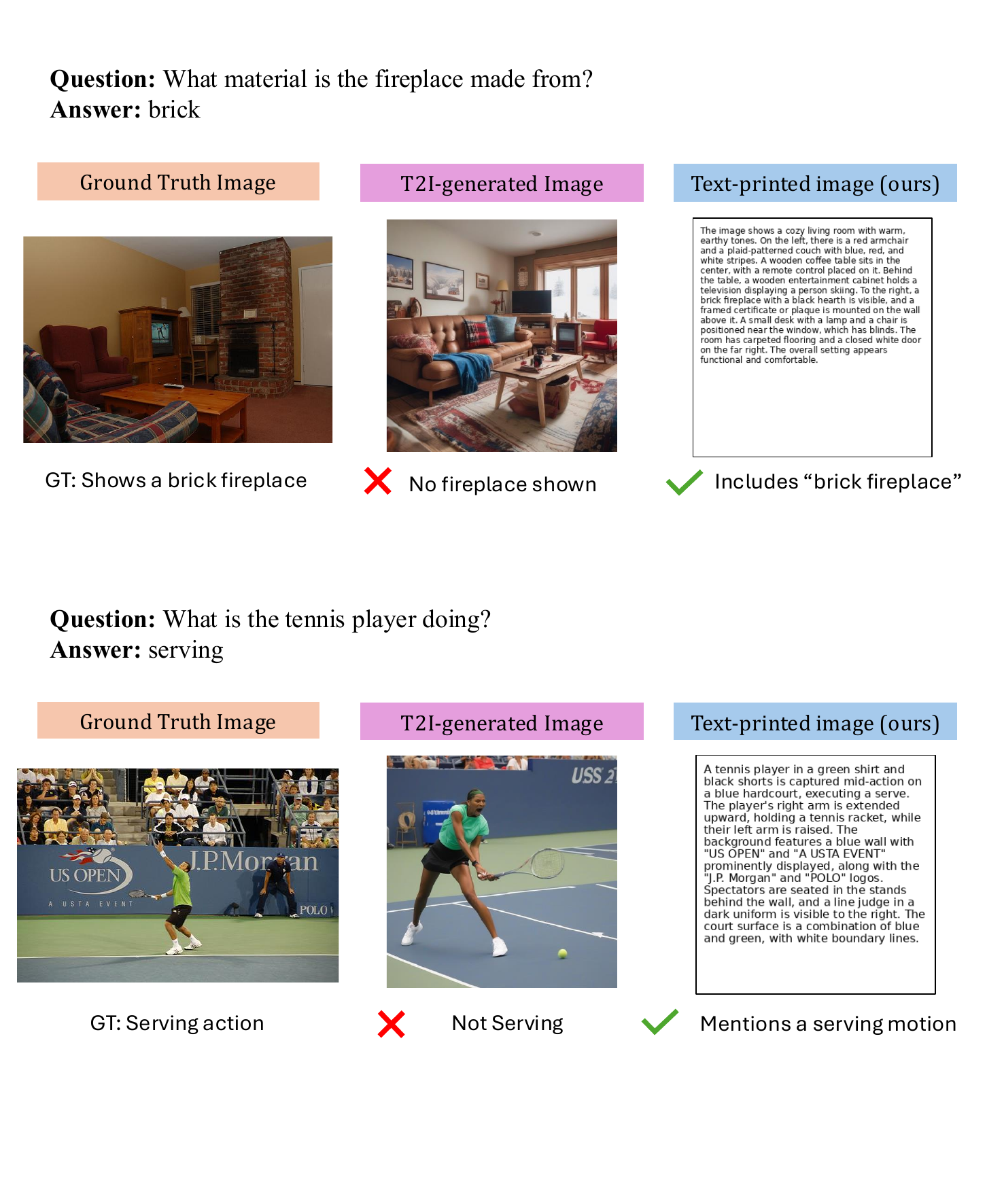}
  \caption{
  Qualitative comparison of synthetic images on OK-VQA.
  }
  \label{fig:qualitative_example_OK-VQA}
\end{figure*}

\begin{figure*}[t]
  \centering
  \includegraphics[width=\linewidth]{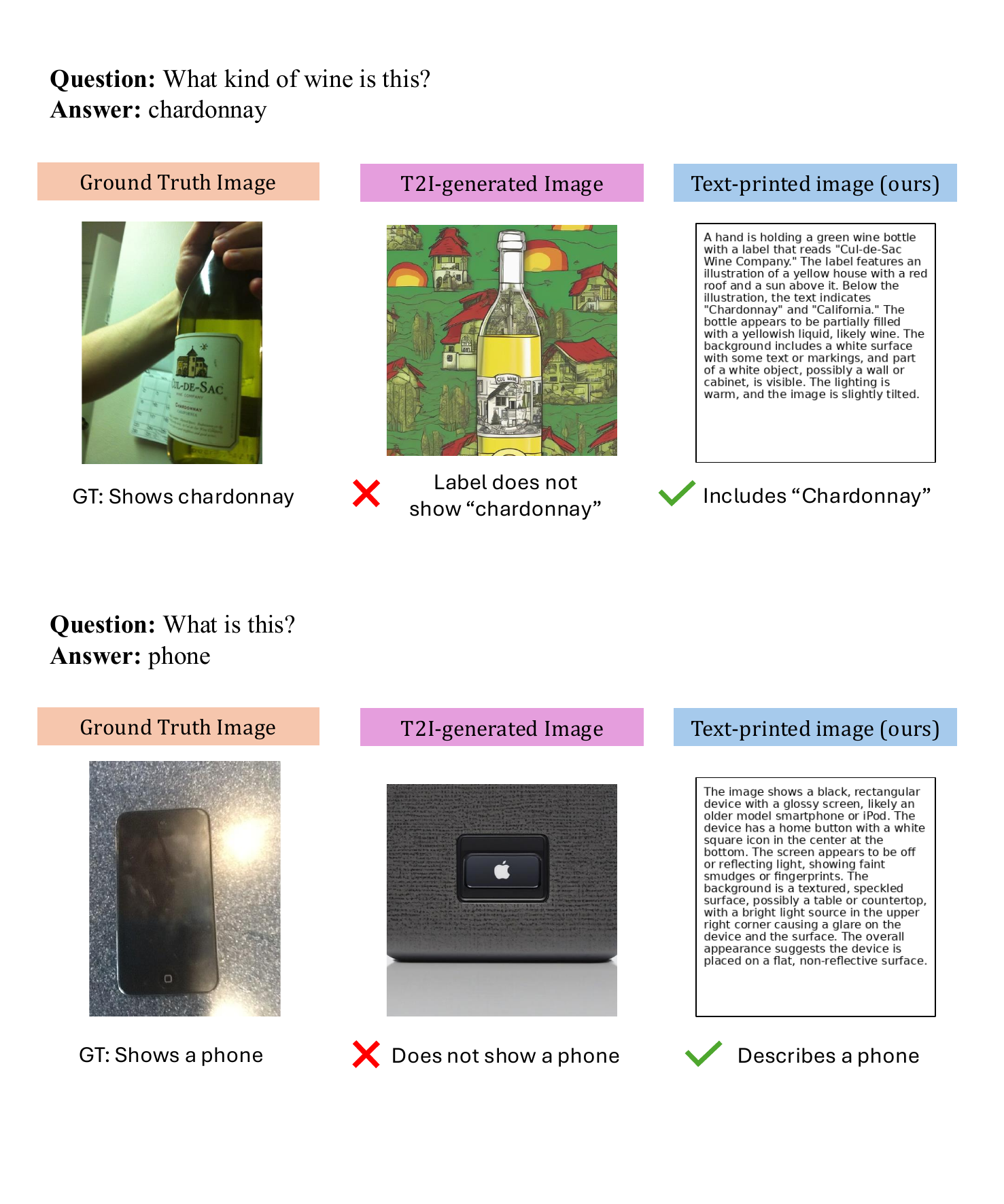}
  \caption{
  Qualitative comparison of synthetic images on VizWiz.
  }
  \label{fig:qualitative_example_vizwiz}
\end{figure*}

\begin{figure*}[t]
  \centering
  \includegraphics[width=\linewidth]{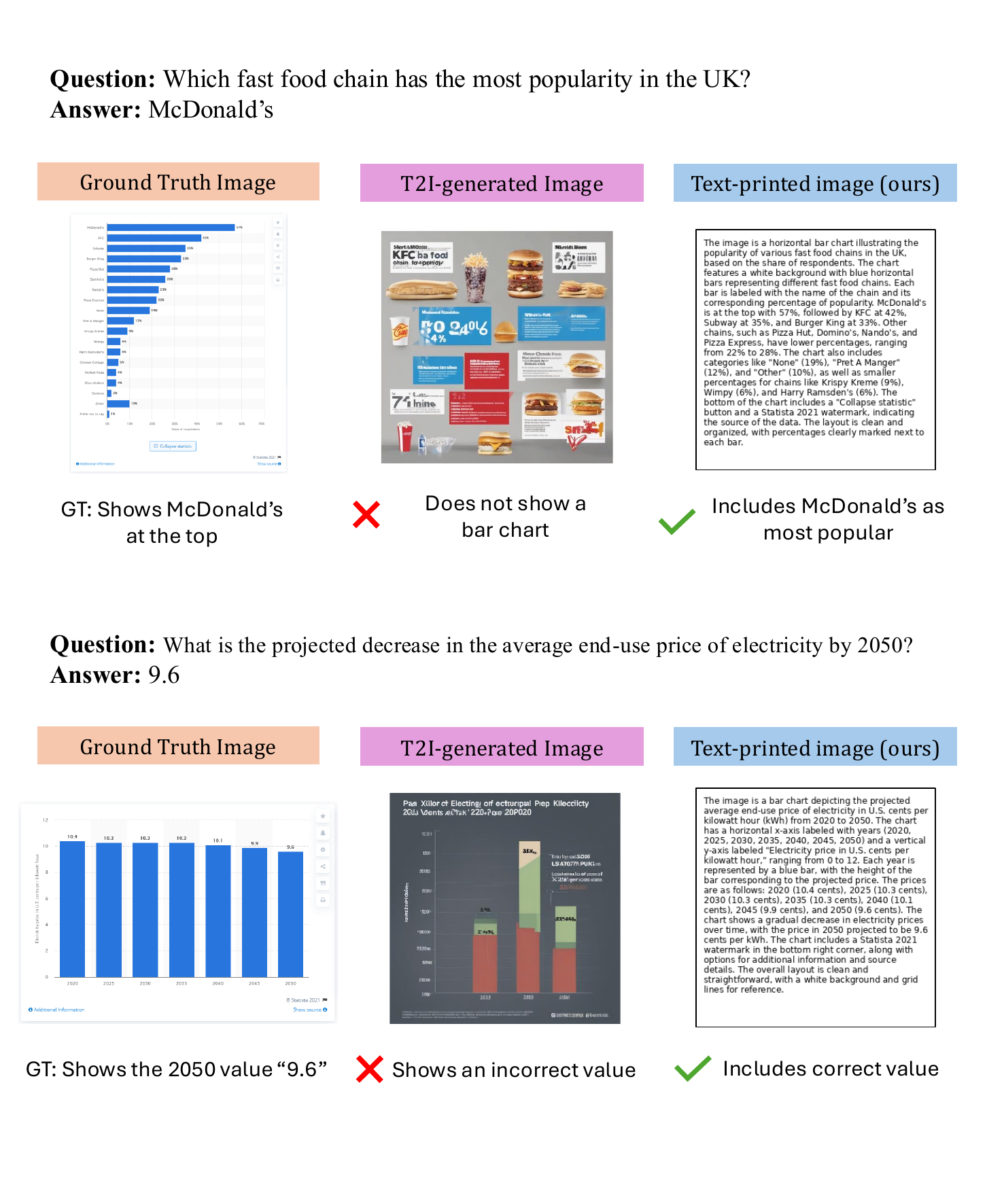}
  \caption{
  Qualitative comparison of synthetic images on ChartQA.
  }
  \label{fig:qualitative_example_chartQA}
\end{figure*}

\begin{figure*}[t]
  \centering
  \includegraphics[width=\linewidth]{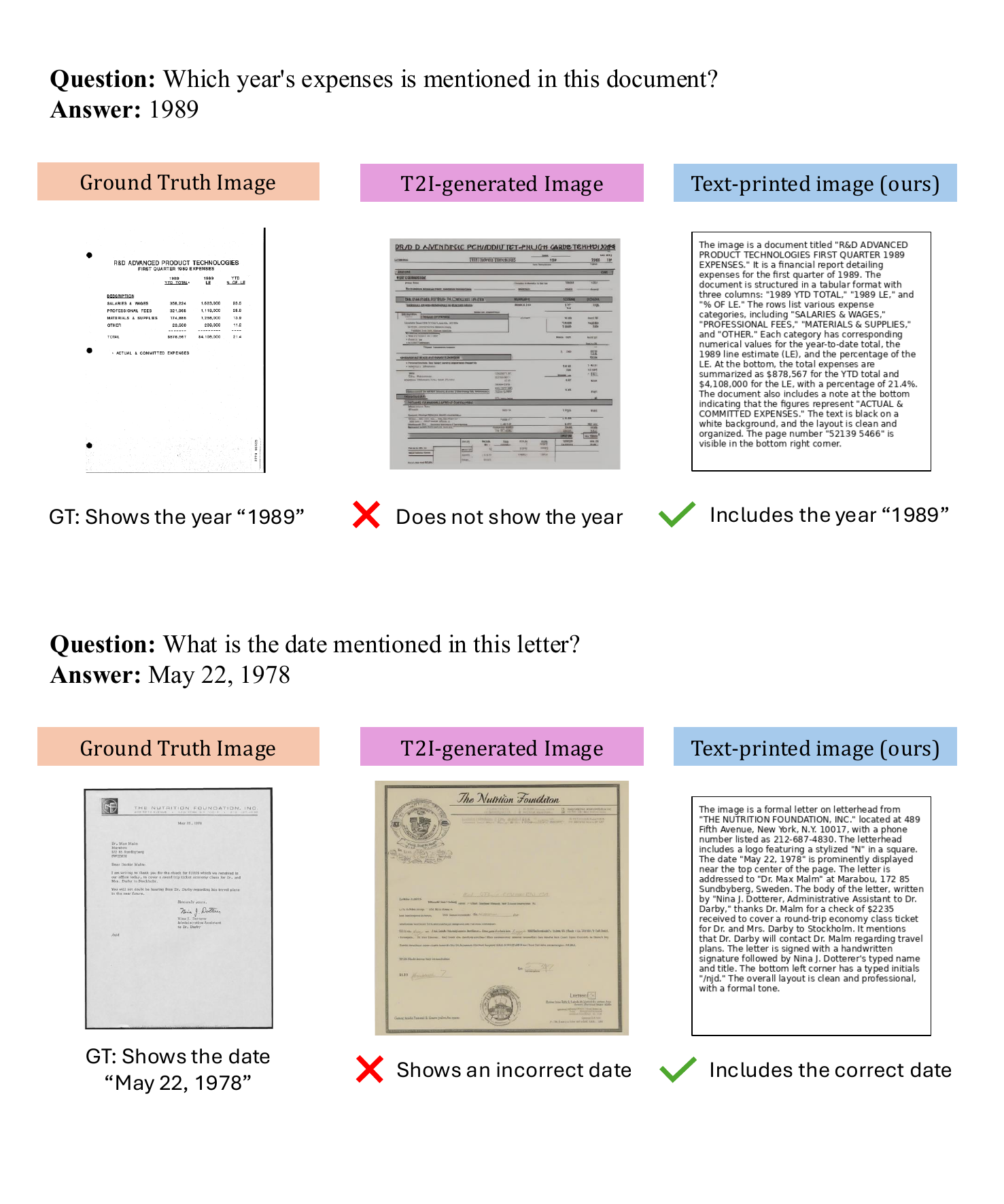}
  \caption{
  Qualitative comparison of synthetic images on DocVQA.
  }
  \label{fig:qualitative_example_docQA}
\end{figure*}

\begin{figure*}[t]
  \centering
  \includegraphics[width=\linewidth]{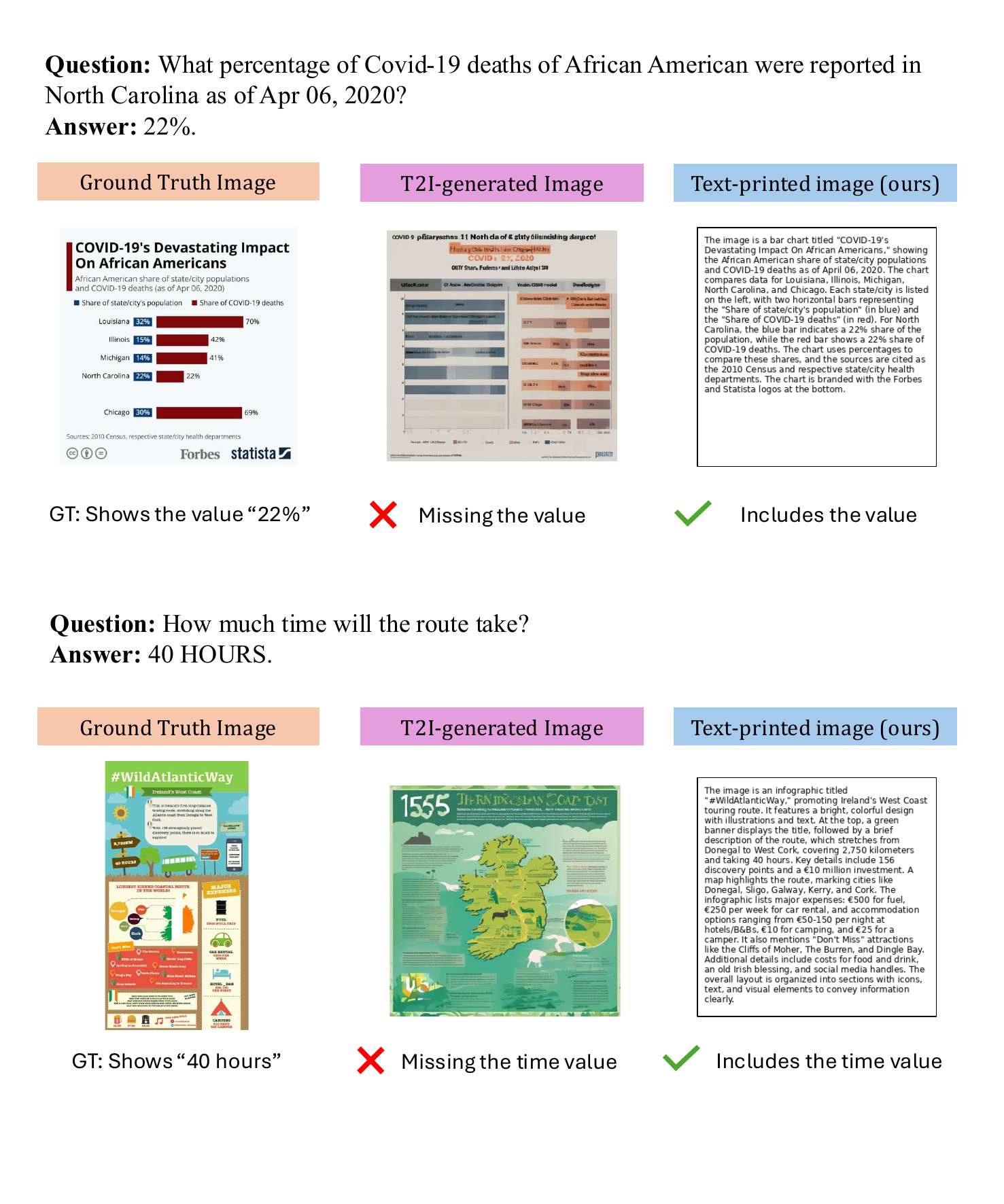}
  \caption{
  Qualitative comparison of synthetic images on InfoVQA.
  }
  \label{fig:qualitative_example_infoQA}
\end{figure*}

Figures~\ref{fig:qualitative_example_scienceQA}--\ref{fig:qualitative_example_infoQA} shows the examples.
They reveal that T2I often produces images that contradict the QA pair or omit essential information. Consistent with our overall findings, T2I struggles especially in Text VQA, where generating readable and accurate text is critical. We believe this difficulty arises from the inherent limitations of T2I models in rendering text reliably.

%% file: main.bib
@String(AAAI = {AAAI})

@article{liu2023visual,
  title={Visual instruction tuning},
  author={Liu, Haotian and Li, Chunyuan and Wu, Qingyang and Lee, Yong Jae},
  journal={Advances in neural information processing systems},
  volume={36},
  pages={34892--34916},
  year={2023}
}

@misc{bai2023qwen,
      title={Qwen-VL: A Versatile Vision-Language Model for Understanding, Localization, Text Reading, and Beyond}, 
      author={Jinze Bai and Shuai Bai and Shusheng Yang and Shijie Wang and Sinan Tan and Peng Wang and Junyang Lin and Chang Zhou and Jingren Zhou},
      year={2023},
      eprint={2308.12966},
      archivePrefix={arXiv},
      primaryClass={cs.CV},
      url={https://arxiv.org/abs/2308.12966}, 
}

@article{wang2024qwen2,
  title={Qwen2-vl: Enhancing vision-language model's perception of the world at any resolution},
  author={Wang, Peng and Bai, Shuai and Tan, Sinan and Wang, Shijie and Fan, Zhihao and Bai, Jinze and Chen, Keqin and Liu, Xuejing and Wang, Jialin and Ge, Wenbin and others},
  journal={arXiv preprint arXiv:2409.12191},
  year={2024}
}

@inproceedings{
li2023decap,
title={DeCap: Decoding {CLIP} Latents for Zero-Shot Captioning via Text-Only Training},
author={Wei Li and Linchao Zhu and Longyin Wen and Yi Yang},
booktitle={The Eleventh International Conference on Learning Representations },
year={2023},
url={https://openreview.net/forum?id=Lt8bMlhiwx2}
}

@article{zhou2022lafite2,
  title={Lafite2: Few-shot text-to-image generation},
  author={Zhou, Yufan and Li, Chunyuan and Chen, Changyou and Gao, Jianfeng and Xu, Jinhui},
  journal={arXiv preprint arXiv:2210.14124},
  year={2022}
}

@inproceedings{nukrai2022text,
  title={Text-Only Training for Image Captioning using Noise-Injected CLIP},
  author={Nukrai, David and Mokady, Ron and Globerson, Amir},
  booktitle={Findings of the Association for Computational Linguistics: EMNLP 2022},
  pages={4055--4063},
  year={2022}
}

@inproceedings{zhou2023shifted,
  title={Shifted diffusion for text-to-image generation},
  author={Zhou, Yufan and Liu, Bingchen and Zhu, Yizhe and Yang, Xiao and Chen, Changyou and Xu, Jinhui},
  booktitle={Proceedings of the IEEE/CVF conference on computer vision and pattern recognition},
  pages={10157--10166},
  year={2023}
}

@inproceedings{
zhang2024connect,
title={Connect, Collapse, Corrupt: Learning Cross-Modal Tasks with Uni-Modal Data},
author={Yuhui Zhang and Elaine Sui and Serena Yeung},
booktitle={The Twelfth International Conference on Learning Representations},
year={2024},
url={https://openreview.net/forum?id=ttXg3SKAg5}
}

@article{yu2025unicorn,
  title={Unicorn: Text-only data synthesis for vision language model training},
  author={Yu, Xiaomin and Ding, Pengxiang and Zhang, Wenjie and Huang, Siteng and Gao, Songyang and Qin, Chengwei and Wu, Kejian and Fan, Zhaoxin and Qiao, Ziyue and Wang, Donglin},
  journal={arXiv preprint arXiv:2503.22655},
  year={2025}
}

@article{hu2025praxis,
  title={Praxis-VLM: Vision-Grounded Decision Making via Text-Driven Reinforcement Learning},
  author={Hu, Zhe and Li, Jing and Pu, Zhongzhu and Chan, Hou Pong and Yin, Yu},
  journal={arXiv preprint arXiv:2503.16965},
  year={2025}
}

@article{choi2024improving,
  title={Improving Fine-grained Visual Understanding in VLMs through Text-Only Training},
  author={Choi, Dasol and Son, Guijin and Kim, Soo Yong and Paik, Gio and Hong, Seunghyeok},
  journal={arXiv preprint arXiv:2412.12940},
  year={2024}
}

@article{li2023llavamed,
  title={Llava-med: Training a large language-and-vision assistant for biomedicine in one day},
  author={Li, Chunyuan and Wong, Cliff and Zhang, Sheng and Usuyama, Naoto and Liu, Haotian and Yang, Jianwei and Naumann, Tristan and Poon, Hoifung and Gao, Jianfeng},
  journal={arXiv preprint arXiv:2306.00890},
  year={2023}
}

@article{zhang2023llavar,
  title={Llavar: Enhanced visual instruction tuning for text-rich image understanding},
  author={Zhang, Yanzhe and Zhang, Ruiyi and Gu, Jiuxiang and Zhou, Yufan and Lipka, Nedim and Yang, Diyi and Sun, Tong},
  journal={arXiv preprint arXiv:2306.17107},
  year={2023}
}

@article{deng2024enhancing,
  title={Enhancing large vision language models with self-training on image comprehension},
  author={Deng, Yihe and Lu, Pan and Yin, Fan and Hu, Ziniu and Shen, Sheng and Gu, Quanquan and Zou, James Y and Chang, Kai-Wei and Wang, Wei},
  journal={Advances in Neural Information Processing Systems},
  volume={37},
  pages={131369--131397},
  year={2024}
}

@inproceedings{wang2024vigc,
  title={Vigc: Visual instruction generation and correction},
  author={Wang, Bin and Wu, Fan and Han, Xiao and Peng, Jiahui and Zhong, Huaping and Zhang, Pan and Dong, Xiaoyi and Li, Weijia and Li, Wei and Wang, Jiaqi and others},
  booktitle={Proceedings of the AAAI Conference on Artificial Intelligence},
  volume={38},
  number={6},
  pages={5309--5317},
  year={2024}
}

@article{dai2025captions,
  title={From captions to rewards (carevl): Leveraging large language model experts for enhanced reward modeling in large vision-language models},
  author={Dai, Muzhi and Sun, Jiashuo and Zhao, Zhiyuan and Liu, Shixuan and Li, Rui and Gao, Junyu and Li, Xuelong},
  journal={arXiv preprint arXiv:2503.06260},
  year={2025}
}

@inproceedings{aboutalebi2024magid,
  title={MAGID: An Automated Pipeline for Generating Synthetic Multi-modal Datasets},
  author={Aboutalebi, Hossein and Song, Hwanjun and Xie, Yusheng and Gupta, Arshit and Sun, Lijia and Su, Hang and Shalyminov, Igor and Pappas, Nikolaos and Singh, Siffi and Mansour, Saab},
  booktitle={Proceedings of the 2024 Conference of the North American Chapter of the Association for Computational Linguistics: Human Language Technologies (Volume 1: Long Papers)},
  pages={5150--5167},
  year={2024}
}

@article{ashrafian2024vision,
  title={Vision-language synthetic data enhances echocardiography downstream tasks},
  author={Ashrafian, Pooria and Yazdani, Milad and Heidari, Moein and Shahriari, Dena and Hacihaliloglu, Ilker},
  journal={arXiv preprint arXiv:2403.19880},
  year={2024}
}

@inproceedings{wu2024autohallusion,
  title={AutoHallusion: Automatic Generation of Hallucination Benchmarks for Vision-Language Models},
  author={Wu, Xiyang and Guan, Tianrui and Li, Dianqi and Huang, Shuaiyi and Liu, Xiaoyu and Wang, Xijun and Xian, Ruiqi and Shrivastava, Abhinav and Huang, Furong and Boyd-Graber, Jordan and others},
  booktitle={Findings of the Association for Computational Linguistics: EMNLP 2024},
  pages={8395--8419},
  year={2024}
}

@inproceedings{ben2024mitigating,
  title={Mitigating Open-Vocabulary Caption Hallucinations},
  author={Ben-Kish, Assaf and Yanuka, Moran and Alper, Morris and Giryes, Raja and Averbuch-Elor, Hadar},
  booktitle={Proceedings of the 2024 Conference on Empirical Methods in Natural Language Processing},
  pages={22680--22698},
  year={2024}
}

@article{wu2025less,
  title={Less-to-more generalization: Unlocking more controllability by in-context generation},
  author={Wu, Shaojin and Huang, Mengqi and Wu, Wenxu and Cheng, Yufeng and Ding, Fei and He, Qian},
  journal={arXiv preprint arXiv:2504.02160},
  year={2025}
}

@inproceedings{lin2024evaluating,
  title={Evaluating text-to-visual generation with image-to-text generation},
  author={Lin, Zhiqiu and Pathak, Deepak and Li, Baiqi and Li, Jiayao and Xia, Xide and Neubig, Graham and Zhang, Pengchuan and Ramanan, Deva},
  booktitle={European Conference on Computer Vision},
  pages={366--384},
  year={2024},
  organization={Springer}
}

@article{zhou2024vision,
  title={Vision language models in autonomous driving: A survey and outlook},
  author={Zhou, Xingcheng and Liu, Mingyu and Yurtsever, Ekim and Zagar, Bare Luka and Zimmer, Walter and Cao, Hu and Knoll, Alois C},
  journal={IEEE Transactions on Intelligent Vehicles},
  year={2024},
  publisher={IEEE}
}

@article{hartsock2024vision,
  title={Vision-language models for medical report generation and visual question answering: A review},
  author={Hartsock, Iryna and Rasool, Ghulam},
  journal={Frontiers in artificial intelligence},
  volume={7},
  pages={1430984},
  year={2024},
  publisher={Frontiers Media SA}
}

@inproceedings{van2024large,
  title={On large visual language models for medical imaging analysis: An empirical study},
  author={Van, Minh--Hao and Verma, Prateek and Wu, Xintao},
  booktitle={2024 IEEE/ACM Conference on Connected Health: Applications, Systems and Engineering Technologies (CHASE)},
  pages={172--176},
  year={2024},
  organization={IEEE}
}

@inproceedings{li2023clip,
  title={Clip-reid: exploiting vision-language model for image re-identification without concrete text labels},
  author={Li, Siyuan and Sun, Li and Li, Qingli},
  booktitle={Proceedings of the AAAI conference on artificial intelligence},
  volume={37},
  number={1},
  pages={1405--1413},
  year={2023}
}

@inproceedings{ding2024data,
  title={Data Augmentation using LLMs: Data Perspectives, Learning Paradigms and Challenges},
  author={Ding, Bosheng and Qin, Chengwei and Zhao, Ruochen and Luo, Tianze and Li, Xinze and Chen, Guizhen and Xia, Wenhan and Hu, Junjie and Luu, Anh Tuan and Joty, Shafiq},
  booktitle={ACL (Findings)},
  year={2024}
}

@inproceedings{kornblith2019similarity,
  title={Similarity of neural network representations revisited},
  author={Kornblith, Simon and Norouzi, Mohammad and Lee, Honglak and Hinton, Geoffrey},
  booktitle={International conference on machine learning},
  pages={3519--3529},
  year={2019},
  organization={PMlR}
}

@article{liu2024ocrbench,
  title={Ocrbench: on the hidden mystery of ocr in large multimodal models},
  author={Liu, Yuliang and Li, Zhang and Huang, Mingxin and Yang, Biao and Yu, Wenwen and Li, Chunyuan and Yin, Xu-Cheng and Liu, Cheng-Lin and Jin, Lianwen and Bai, Xiang},
  journal={Science China Information Sciences},
  volume={67},
  number={12},
  pages={220102},
  year={2024},
  publisher={Springer}
}

@inproceedings{singh2019towards,
  title={Towards vqa models that can read},
  author={Singh, Amanpreet and Natarajan, Vivek and Shah, Meet and Jiang, Yu and Chen, Xinlei and Batra, Dhruv and Parikh, Devi and Rohrbach, Marcus},
  booktitle={Proceedings of the IEEE/CVF conference on computer vision and pattern recognition},
  pages={8317--8326},
  year={2019}
}

@article{dai2023instructblip,
  title={Instructblip: Towards general-purpose vision-language models with instruction tuning},
  author={Dai, Wenliang and Li, Junnan and Li, Dongxu and Tiong, Anthony and Zhao, Junqi and Wang, Weisheng and Li, Boyang and Fung, Pascale N and Hoi, Steven},
  journal={Advances in neural information processing systems},
  volume={36},
  pages={49250--49267},
  year={2023}
}

@inproceedings{
zhu2024minigpt,
title={Mini{GPT}-4: Enhancing Vision-Language Understanding with Advanced Large Language Models},
author={Deyao Zhu and Jun Chen and Xiaoqian Shen and Xiang Li and Mohamed Elhoseiny},
booktitle={The Twelfth International Conference on Learning Representations},
year={2024},
url={https://openreview.net/forum?id=1tZbq88f27}
}

@inproceedings{lin2024vila,
  title={Vila: On pre-training for visual language models},
  author={Lin, Ji and Yin, Hongxu and Ping, Wei and Molchanov, Pavlo and Shoeybi, Mohammad and Han, Song},
  booktitle={Proceedings of the IEEE/CVF conference on computer vision and pattern recognition},
  pages={26689--26699},
  year={2024}
}

@inproceedings{chen2024internvl,
  title={Internvl: Scaling up vision foundation models and aligning for generic visual-linguistic tasks},
  author={Chen, Zhe and Wu, Jiannan and Wang, Wenhai and Su, Weijie and Chen, Guo and Xing, Sen and Zhong, Muyan and Zhang, Qinglong and Zhu, Xizhou and Lu, Lewei and others},
  booktitle={Proceedings of the IEEE/CVF conference on computer vision and pattern recognition},
  pages={24185--24198},
  year={2024}
}

@article{tong2024cambrian,
  title={Cambrian-1: A fully open, vision-centric exploration of multimodal llms},
  author={Tong, Peter and Brown, Ellis and Wu, Penghao and Woo, Sanghyun and IYER, Adithya Jairam Vedagiri and Akula, Sai Charitha and Yang, Shusheng and Yang, Jihan and Middepogu, Manoj and Wang, Ziteng and others},
  journal={Advances in Neural Information Processing Systems},
  volume={37},
  pages={87310--87356},
  year={2024}
}

@article{wang2025internvl3,
  title={Internvl3. 5: Advancing open-source multimodal models in versatility, reasoning, and efficiency},
  author={Wang, Weiyun and Gao, Zhangwei and Gu, Lixin and Pu, Hengjun and Cui, Long and Wei, Xingguang and Liu, Zhaoyang and Jing, Linglin and Ye, Shenglong and Shao, Jie and others},
  journal={arXiv preprint arXiv:2508.18265},
  year={2025}
}

@article{zhu2025internvl3,
  title={Internvl3: Exploring advanced training and test-time recipes for open-source multimodal models},
  author={Zhu, Jinguo and Wang, Weiyun and Chen, Zhe and Liu, Zhaoyang and Ye, Shenglong and Gu, Lixin and Tian, Hao and Duan, Yuchen and Su, Weijie and Shao, Jie and others},
  journal={arXiv preprint arXiv:2504.10479},
  year={2025}
}

@article{dubey2024llama,
  title={The llama 3 herd of models},
  author={Dubey, Abhimanyu and Jauhri, Abhinav and Pandey, Abhinav and Kadian, Abhishek and Al-Dahle, Ahmad and Letman, Aiesha and Mathur, Akhil and Schelten, Alan and Yang, Amy and Fan, Angela and others},
  journal={arXiv e-prints},
  pages={arXiv--2407},
  year={2024}
}

@inproceedings{li2024enhanced,
  title={Enhanced Visual Instruction Tuning with Synthesized Image-Dialogue Data},
  author={Li, Yanda and Zhang, Chi and Yu, Gang and Yang, Wanqi and Wang, Zhibin and Fu, Bin and Lin, Guosheng and Shen, Chunhua and Chen, Ling and Wei, Yunchao},
  booktitle={Findings of the Association for Computational Linguistics ACL 2024},
  pages={14512--14531},
  year={2024}
}

@article{sharifzadeh2024synth,
  title={Synth $^{2}$: Boosting Visual-Language Models with Synthetic Captions and Image Embeddings},
  author={Sharifzadeh, Sahand and Kaplanis, Christos and Pathak, Shreya and Kumaran, Dharshan and Ilic, Anastasija and Mitrovic, Jovana and Blundell, Charles and Banino, Andrea},
  journal={arXiv preprint arXiv:2403.07750},
  year={2024}
}

@inproceedings{wang2023self,
  title={Self-Instruct: Aligning Language Models with Self-Generated Instructions},
  author={Wang, Yizhong and Kordi, Yeganeh and Mishra, Swaroop and Liu, Alisa and Smith, Noah A and Khashabi, Daniel and Hajishirzi, Hannaneh},
  booktitle={Proceedings of the 61st Annual Meeting of the Association for Computational Linguistics (Volume 1: Long Papers)},
  pages={13484--13508},
  year={2023}
}

@inproceedings{marino2019ok,
  title={Ok-vqa: A visual question answering benchmark requiring external knowledge},
  author={Marino, Kenneth and Rastegari, Mohammad and Farhadi, Ali and Mottaghi, Roozbeh},
  booktitle={Proceedings of the IEEE/cvf conference on computer vision and pattern recognition},
  pages={3195--3204},
  year={2019}
}

@inproceedings{gurari2018vizwiz,
  title={Vizwiz grand challenge: Answering visual questions from blind people},
  author={Gurari, Danna and Li, Qing and Stangl, Abigale J and Guo, Anhong and Lin, Chi and Grauman, Kristen and Luo, Jiebo and Bigham, Jeffrey P},
  booktitle={Proceedings of the IEEE conference on computer vision and pattern recognition},
  pages={3608--3617},
  year={2018}
}

@inproceedings{mathew2021docvqa,
  title={Docvqa: A dataset for vqa on document images},
  author={Mathew, Minesh and Karatzas, Dimosthenis and Jawahar, CV},
  booktitle={Proceedings of the IEEE/CVF winter conference on applications of computer vision},
  pages={2200--2209},
  year={2021}
}

@inproceedings{mathew2022infographicvqa,
  title={Infographicvqa},
  author={Mathew, Minesh and Bagal, Viraj and Tito, Rub{\`e}n and Karatzas, Dimosthenis and Valveny, Ernest and Jawahar, CV},
  booktitle={Proceedings of the IEEE/CVF Winter Conference on Applications of Computer Vision},
  pages={1697--1706},
  year={2022}
}

@article{masry2022chartqa,
  title={Chartqa: A benchmark for question answering about charts with visual and logical reasoning},
  author={Masry, Ahmed and Long, Do Xuan and Tan, Jia Qing and Joty, Shafiq and Hoque, Enamul},
  journal={arXiv preprint arXiv:2203.10244},
  year={2022}
}

@inproceedings{sima2024drivelm,
  title={Drivelm: Driving with graph visual question answering},
  author={Sima, Chonghao and Renz, Katrin and Chitta, Kashyap and Chen, Li and Zhang, Hanxue and Xie, Chengen and Bei{\ss}wenger, Jens and Luo, Ping and Geiger, Andreas and Li, Hongyang},
  booktitle={European conference on computer vision},
  pages={256--274},
  year={2024},
  organization={Springer}
}

@misc{zhang2024lmmsevalrealitycheckevaluation,
      title={LMMs-Eval: Reality Check on the Evaluation of Large Multimodal Models}, 
      author={Kaichen Zhang and Bo Li and Peiyuan Zhang and Fanyi Pu and Joshua Adrian Cahyono and Kairui Hu and Shuai Liu and Yuanhan Zhang and Jingkang Yang and Chunyuan Li and Ziwei Liu},
      year={2024},
      eprint={2407.12772},
      archivePrefix={arXiv},
      primaryClass={cs.CL},
      url={https://arxiv.org/abs/2407.12772}, 
}

@inproceedings{
podell2024sdxl,
title={{SDXL}: Improving Latent Diffusion Models for High-Resolution Image Synthesis},
author={Dustin Podell and Zion English and Kyle Lacey and Andreas Blattmann and Tim Dockhorn and Jonas M{\"u}ller and Joe Penna and Robin Rombach},
booktitle={The Twelfth International Conference on Learning Representations},
year={2024},
url={https://openreview.net/forum?id=di52zR8xgf}
}

@inproceedings{
garrido2023on,
title={On the duality between contrastive and non-contrastive self-supervised learning},
author={Quentin Garrido and Yubei Chen and Adrien Bardes and Laurent Najman and Yann LeCun},
booktitle={The Eleventh International Conference on Learning Representations },
year={2023},
url={https://openreview.net/forum?id=kDEL91Dufpa}
}

@article{huang2023t2i,
  title={T2i-compbench: A comprehensive benchmark for open-world compositional text-to-image generation},
  author={Huang, Kaiyi and Sun, Kaiyue and Xie, Enze and Li, Zhenguo and Liu, Xihui},
  journal={Advances in Neural Information Processing Systems},
  volume={36},
  pages={78723--78747},
  year={2023}
}

@inproceedings{hu2023tifa,
  title={Tifa: Accurate and interpretable text-to-image faithfulness evaluation with question answering},
  author={Hu, Yushi and Liu, Benlin and Kasai, Jungo and Wang, Yizhong and Ostendorf, Mari and Krishna, Ranjay and Smith, Noah A},
  booktitle={Proceedings of the IEEE/CVF International Conference on Computer Vision},
  pages={20406--20417},
  year={2023}
}

@article{liang2022mind,
  title={Mind the gap: Understanding the modality gap in multi-modal contrastive representation learning},
  author={Liang, Victor Weixin and Zhang, Yuhui and Kwon, Yongchan and Yeung, Serena and Zou, James Y},
  journal={Advances in Neural Information Processing Systems},
  volume={35},
  pages={17612--17625},
  year={2022}
}

@inproceedings{
zhang2023diagnosing,
title={Diagnosing and Rectifying Vision Models using Language},
author={Yuhui Zhang and Jeff Z. HaoChen and Shih-Cheng Huang and Kuan-Chieh Wang and James Zou and Serena Yeung},
booktitle={The Eleventh International Conference on Learning Representations },
year={2023},
url={https://openreview.net/forum?id=D-zfUK7BR6c}
}

@article{du2025virgo,
  title={Virgo: A preliminary exploration on reproducing o1-like mllm},
  author={Du, Yifan and Liu, Zikang and Li, Yifan and Zhao, Wayne Xin and Huo, Yuqi and Wang, Bingning and Chen, Weipeng and Liu, Zheng and Wang, Zhongyuan and Wen, Ji-Rong},
  journal={arXiv preprint arXiv:2501.01904},
  year={2025}
}

@inproceedings{zhao2025omnialign,
  title={Omnialign-v: Towards enhanced alignment of mllms with human preference},
  author={Zhao, Xiangyu and Ding, Shengyuan and Zhang, Zicheng and Huang, Haian and Maosongcao, Maosongcao and Wang, Jiaqi and Wang, Weiyun and Fang, Xinyu and Wang, Wenhai and Zhai, Guangtao and others},
  booktitle={Proceedings of the 63rd Annual Meeting of the Association for Computational Linguistics (Volume 1: Long Papers)},
  pages={18490--18515},
  year={2025}
}

@article{chen2024allava,
  title={Allava: Harnessing gpt4v-synthesized data for lite vision-language models},
  author={Chen, Guiming Hardy and Chen, Shunian and Zhang, Ruifei and Chen, Junying and Wu, Xiangbo and Zhang, Zhiyi and Chen, Zhihong and Li, Jianquan and Wan, Xiang and Wang, Benyou},
  journal={arXiv preprint arXiv:2402.11684},
  year={2024}
}

@inproceedings{zhao2024genixer,
  title={GENIXER: Empowering Multimodal Large Language Model as a Powerful Data Generator},
  author={Zhao, Henry Hengyuan and Zhou, Pan and Shou, Mike Zheng},
  booktitle={European Conference on Computer Vision},
  pages={129--147},
  year={2024}
}

@misc{
gui2025cycleaug,
title={CycleAug: Cycle-Consistent Visual Augmentation for Large Multimodal Models},
author={Zhongrui Gui and Luoxin Ye and Wufei Ma and Zhao-Yang Wang and Ariel Lubonja and Daniel Khashabi and Alan Yuille and Jieneng Chen},
year={2025},
url={https://openreview.net/forum?id=IiRlImvLQI}
}

@inproceedings{irawan2025towards,
  title={Towards efficient and robust vqa-nle data generation with large vision-language models},
  author={Irawan, Patrick Amadeus and Winata, Genta Indra and Cahyawijaya, Samuel and Purwarianti, Ayu},
  booktitle={Proceedings of the 31st International Conference on Computational Linguistics},
  pages={4323--4340},
  year={2025}
}

@inproceedings{
trabucco2024effective,
title={Effective Data Augmentation With Diffusion Models},
author={Brandon Trabucco and Kyle Doherty and Max A Gurinas and Ruslan Salakhutdinov},
booktitle={The Twelfth International Conference on Learning Representations},
year={2024},
url={https://openreview.net/forum?id=ZWzUA9zeAg}
}

@article{wang2025endochat,
  title={Endochat: Grounded multimodal large language model for endoscopic surgery},
  author={Wang, Guankun and Bai, Long and Wang, Junyi and Yuan, Kun and Li, Zhen and Jiang, Tianxu and He, Xiting and Wu, Jinlin and Chen, Zhen and Lei, Zhen and others},
  journal={arXiv preprint arXiv:2501.11347},
  year={2025}
}

@inproceedings{
zhou2025anyprefer,
title={Anyprefer: An Agentic Framework for Preference Data Synthesis},
author={Yiyang Zhou and Zhaoyang Wang and Tianle Wang and Shangyu Xing and Peng Xia and Bo Li and Kaiyuan Zheng and Zijian Zhang and Zhaorun Chen and Wenhao Zheng and Xuchao Zhang and Chetan Bansal and Weitong Zhang and Ying Wei and Mohit Bansal and Huaxiu Yao},
booktitle={The Thirteenth International Conference on Learning Representations},
year={2025},
url={https://openreview.net/forum?id=WpZyPk79Fu}
}

@inproceedings{lee2021constructing,
  title={Constructing multi-modal dialogue dataset by replacing text with semantically relevant images},
  author={Lee, Nyoungwoo and Shin, Suwon and Choo, Jaegul and Choi, Ho-Jin and Myaeng, Sung-Hyon},
  booktitle={Proceedings of the 59th Annual Meeting of the Association for Computational Linguistics and the 11th International Joint Conference on Natural Language Processing (Volume 2: Short Papers)},
  pages={897--906},
  year={2021}
}

@inproceedings{lee2024dialogcc,
  title={Dialogcc: An automated pipeline for creating high-quality multi-modal dialogue dataset},
  author={Lee, Young-Jun and Ko, Byungsoo and Kim, Han-Gyu and Hyeon, Jonghwan and Choi, Ho-Jin},
  booktitle={Proceedings of the 2024 Conference of the North American Chapter of the Association for Computational Linguistics: Human Language Technologies (Volume 1: Long Papers)},
  pages={1938--1963},
  year={2024}
}

@inproceedings{wu2025synthetic,
  title={Synthetic Data is an  GIFT for Continual Vision-Language Models},
  author={Wu, Bin and Shi, Wuxuan and Wang, Jinqiao and Ye, Mang},
  booktitle={Proceedings of the Computer Vision and Pattern Recognition Conference},
  pages={2813--2823},
  year={2025}
}

@inproceedings{fan2024scaling,
  title={Scaling laws of synthetic images for model training... for now},
  author={Fan, Lijie and Chen, Kaifeng and Krishnan, Dilip and Katabi, Dina and Isola, Phillip and Tian, Yonglong},
  booktitle={Proceedings of the IEEE/CVF Conference on Computer Vision and Pattern Recognition},
  pages={7382--7392},
  year={2024}
}

@article{xu2024vision,
  title={Vision-flan: Scaling human-labeled tasks in visual instruction tuning},
  author={Xu, Zhiyang and Feng, Chao and Shao, Rulin and Ashby, Trevor and Shen, Ying and Jin, Di and Cheng, Yu and Wang, Qifan and Huang, Lifu},
  journal={arXiv preprint arXiv:2402.11690},
  year={2024}
}

@inproceedings{xu2023multiinstruct,
  title={Multiinstruct: Improving multi-modal zero-shot learning via instruction tuning},
  author={Xu, Zhiyang and Shen, Ying and Huang, Lifu},
  booktitle={Proceedings of the 61st Annual Meeting of the Association for Computational Linguistics (Volume 1: Long Papers)},
  pages={11445--11465},
  year={2023}
}

@article{hammoud2024synthclip,
  title={SynthCLIP: Are we ready for a fully synthetic CLIP training?},
  author={Hammoud, Hasan Abed Al Kader and Itani, Hani and Pizzati, Fabio and Torr, Philip and Bibi, Adel and Ghanem, Bernard},
  journal={arXiv preprint arXiv:2402.01832},
  year={2024}
}

@misc{clark2015pillow,
  title={Pillow (PIL Fork) Documentation},
  author={Clark, Alex},
  year={2015},
  publisher={readthedocs},
 url={https://buildmedia.readthedocs.org/media/pdf/pillow/latest/pillow.pdf}
}

@inproceedings{
yaras2025explaining,
title={Explaining and Mitigating the Modality Gap in Contrastive Multimodal Learning},
author={Can Yaras and Siyi Chen and Peng Wang and Qing Qu},
booktitle={The Second Conference on Parsimony and Learning (Proceedings Track)},
year={2025},
url={https://openreview.net/forum?id=2sThreW73a}
}

@inproceedings{zhou2024empirical,
  title={An empirical study on parameter-efficient fine-tuning for multimodal large language models},
  author={Zhou, Xiongtao and He, Jie and Ke, Yuhua and Zhu, Guangyao and Guti{\'e}rrez-Basulto, V{\'\i}ctor and Pan, Jeff},
  booktitle={Findings of the Association for Computational Linguistics: ACL 2024},
  pages={10057--10084},
  year={2024}
}

@article{brown2020language,
  title={Language models are few-shot learners},
  author={Brown, Tom and Mann, Benjamin and Ryder, Nick and Subbiah, Melanie and Kaplan, Jared D and Dhariwal, Prafulla and Neelakantan, Arvind and Shyam, Pranav and Sastry, Girish and Askell, Amanda and others},
  journal={Advances in neural information processing systems},
  volume={33},
  pages={1877--1901},
  year={2020}
}

@article{
li2024visionlanguage,
title={Vision-Language Instruction Tuning: A Review and Analysis},
author={Chen Li and Yixiao Ge and Dian Li and Ying Shan},
journal={Transactions on Machine Learning Research},
issn={2835-8856},
year={2024},
url={https://openreview.net/forum?id=ul2tbUPtIQ},
note={Survey Certification}
}

@inproceedings{wei2019eda,
  title={EDA: Easy Data Augmentation Techniques for Boosting Performance on Text Classification Tasks},
  author={Wei, Jason and Zou, Kai},
  booktitle={Proceedings of the 2019 Conference on Empirical Methods in Natural Language Processing and the 9th International Joint Conference on Natural Language Processing (EMNLP-IJCNLP)},
  pages={6382--6388},
  year={2019}
}

@inproceedings{
hertz2023prompttoprompt,
title={Prompt-to-Prompt Image Editing with Cross-Attention Control},
author={Amir Hertz and Ron Mokady and Jay Tenenbaum and Kfir Aberman and Yael Pritch and Daniel Cohen-Or},
booktitle={The Eleventh International Conference on Learning Representations },
year={2023},
url={https://openreview.net/forum?id=_CDixzkzeyb}
}
